\documentclass[10pt,twocolumn,letterpaper]{article}
\usepackage{cvpr}
\usepackage{times}
\usepackage{epsfig}
\usepackage{graphicx}
\usepackage{amsmath}
\usepackage{amssymb}
\usepackage{arydshln}
\usepackage{adjustbox}
\usepackage{multirow}
\usepackage{flafter}
\usepackage[skip=2pt,font=small]{caption}
\usepackage{subcaption}

\newcommand{\Fig}{Fig }
\newcommand{\Tab}{Table }

\newcommand{\Sec}{Sec }

\newcommand{\aar}{}
\newcommand{\prg}[1]{\par\noindent{{\textbf{#1}}}}

\newcommand{\argmax}{\mathop{\rm arg~max}\limits}

\usepackage[pagebackref=true,breaklinks=true,letterpaper=true,colorlinks,bookmarks=false]{hyperref}

\cvprfinalcopy 

\def\cvprPaperID{654} 

\ifcvprfinal\fi
\begin{document}

\title{Tabulated MLP for Fast Point Feature Embedding}
\author{Yusuke Sekikawa and Teppei Suzuki\\
DENSO IT Laboratory\\
{\tt\small \{ysekikawa,tsuzuki\}@d-itlab.co.jp}
}

\maketitle
\begin{abstract}
Aiming at a drastic speedup for point-data embeddings at test time, we propose a new framework that uses a pair of multi-layer perceptron (MLP) and look-up table (LUT) to transform point-coordinate inputs into high-dimensional features.
When compared with PointNet’s \cite{qi2017pointnet} feature embedding part realized by MLP that requires millions of dot products, ours at test time requires no such layers of matrix-vector products but requires only looking up the nearest entities followed by interpolation, from the tabulated MLP defined over discrete inputs on a 3D lattice.
We call this framework as "LUTI-MLP: LUT Interpolation MLP" that provides a way to train end-to-end tabulated MLP coupled to a LUT in a specific manner without the need for any approximation at test time.
LUTI-MLP also provides significant speedup for Jacobian computation of the embedding function wrt global pose coordinate on Lie algebra $\mathfrak{se}(3)$ at test time, which could be used for point-set registration problems. 
After extensive architectural analysis using ModelNet40 dataset \cite{wu20153d}, we confirmed that our LUTI-MLP even with a small-sized table ($8\times 8\times 8$) yields performance comparable to that of MLP while achieving significant speedup: $80\times$ for embedding, $12\times$ for approximate Jacobian, and $860\times$ for canonical Jacobian. 
\end{abstract}

\begin{figure*}[h!]
\begin{center}
\includegraphics[width=2\columnwidth]{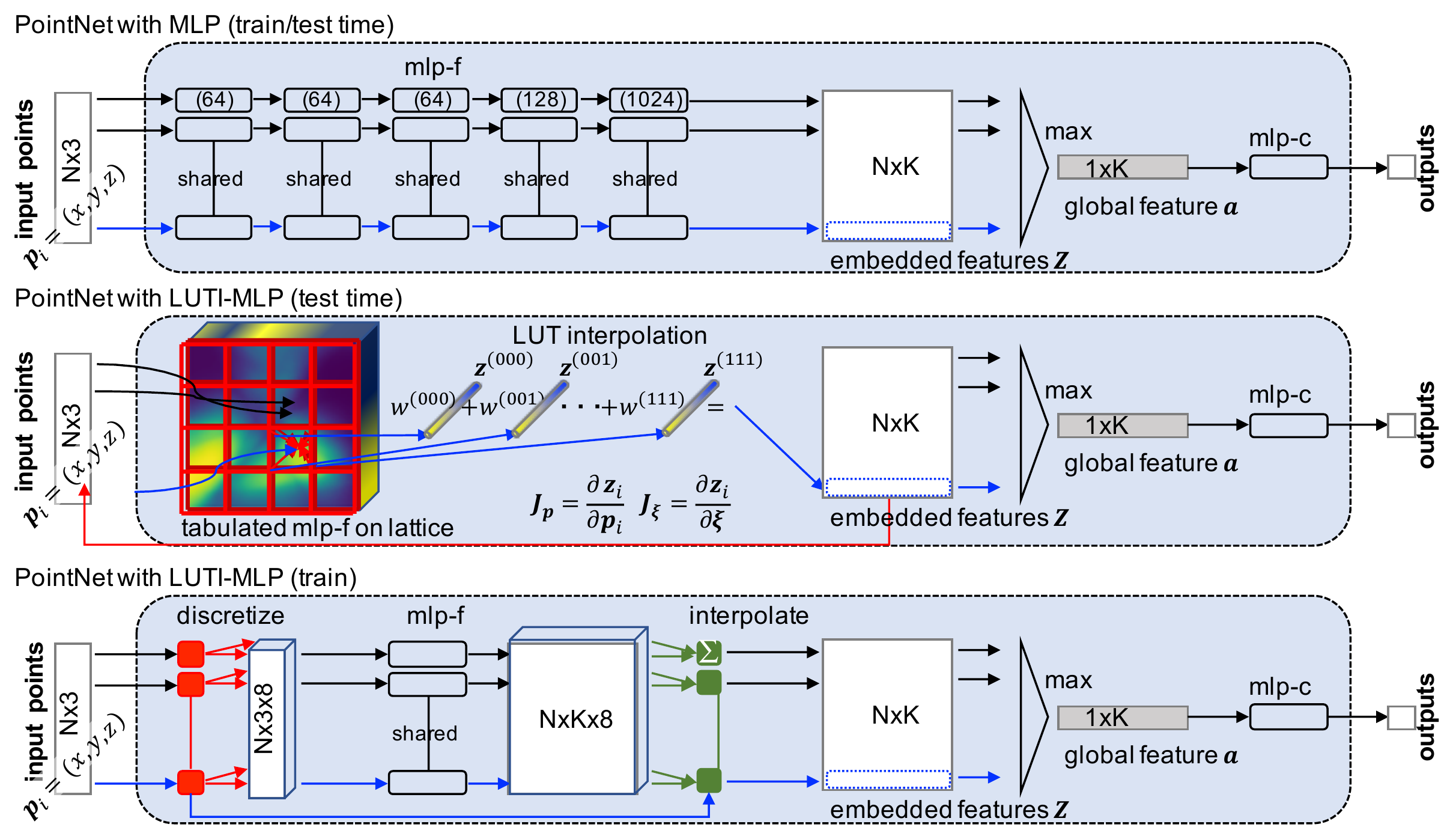}
\end{center}
\caption{
\label{fig:overview}
\textbf{PointNet with our LUTI-MLP compared with the vanilla PointNet with MLP.}
Top: Original PointNet for an object classification network using MLP.
Middle: PointNet by using our LUTI-MLP at test time.
Bottom: PointNet by using our LUTI-MLP at training.
The network structure of LUTI-MLP is asymmetric at train and test time. 
At test time, 8 embedding basis function $\mathbf{\bar{z}}_i^{(j)}$ pre-computed from 8 discrete neighborhood points on the input space lattice $\mathbb{D}^3$ is linearly interpolated (trilinear interpolation) to compute the embedding vector for input $\mathbf{p}_i=[x,y,z]$ in $\mathbb{R}^3$, 
as  $\mathbf{z}_{i}=\sum w_i^{(j)}\mathbf{\bar{z}}_i^{(j)}$.
LUTI-MLP significantly speeds up the embedding by eliminating the vector-matrix multiplication of MLP.
Furthermore, the Jacobian of the embedding vector wrt $\mathbb{R}^3$ and $\mathfrak{se}(3)$, can also be computed efficiently (red line) for test time optimization.
At training, the basis function $\mathbf{\bar{z}}_i^{(j)}= \phi_D(\mathbf{\bar{p}})$ is trained as MLP in end-to-end manner in a same way as at test time.
T-Net is omitted for clarity.
}
\end{figure*}

\section{Introduction}
Point sets are collections of points in a Euclidean space, such as a geometric 3D point cloud captured by depth sensors.
As a set, such data have to be invariant to permutations of its set members (e.g., the point cloud of a bunny is still a bunny, regardless of the order of each point in the set). 

PointNet \cite{qi2017pointnet} is a pioneering work that realized the direct processing of point sets, using point-wise point feature embeddings followed by aggregation with symmetric function. 
Because of its simplicity and performance, the basic idea of the point feature embedding is utilized and extend by wide range of researches \cite{qi2017pointnet,qi2017pointnetplusplus,zhou2018voxelnet,qi2018frustum, lang2019pointpillars,aoki2019pointnetlk,Sarode2019PCRNetPC, lu19,wang2019deep,gross2019alignnet}.
The embedding of each point coordinate is realized with multi-layer perceptron (MLP) that process thousands of input point in the set.
MLP consists of layers of matrix-vector product operation followed by a nonlinearity. 
This requires millions of dot products to process just one point.
For these reasons, the computation of the embedding dominates most of the process.

To this end, we propose a novel framework that compute the embedding using a linear combination of basis functions stored in a lookup table (LUT), which we call \textit{LUT interpolation MLP (LUTI-MLP)}, that is significantly more efficient than MLP. 
It can replace any point feature embedding MLP used in PointNet-like architectures for speedup.
For an input in $\mathbb{R}^3$ such as 3D point cloud, LUTI-MLP computes the embedding of the continuous input point using a linear combination of basis functions evaluated by MLP defined over discrete inputs on a 3D lattice  $\mathbb{D}^3\triangleq D\times D\times D$.
This means, in LUTI-MLP at test time, the embedding no-longer depends on the input point through MLP, rather it depends through the weights of the basis functions, where the basis function are pre-computable as LUT.

\prg{Fast Embedding}
The use of LUT drastically speeds up the embedding process by avoiding the computationally intensive matrix-vector product operation at test time.
With our LUTI-MLP, the embedding computation at test time requires only looking up the nearest entities from LUT  followed by interpolation using the proximity of input to lattice $\mathbb{D}^3$ as weights (\Fig\ref{fig:overview}, middle, \textit{blue} line).

\prg{Fast Jacobian wrt $\mathfrak{se}(3)$}
The difference of aggregated embedding from source and target point sets are used in point-set registration problems \cite{aoki2019pointnetlk}.
In this case, Jacobian of embedding wrt global pose coordinate (Lie algebra $\mathfrak{se}(3)$) needs to be computed online at test time.
The local independence of the embedding on MLP also provides direct differential operation of the embedding wrt $\mathbb{R}^3$ that can be pulled back to the differential wrt $\mathfrak{se}(3)$ (\Fig\ref{fig:overview}, middle, \textit{red} line).
 
\prg{Experiments}
With an extensive evaluation on the ModelNet40 \cite{wu20153d} object classification task, we confirmed that PointNet using our LUTI-MLP even with a small size table ($8\times8\times8$) achieves slightly better performance ($88.01$\% for $D=64$, $87.74$ for $D=8$) than the original ($87.03$\%) while achieving more than $80\times$ speedup for the embedding. 
The proposed LUTI-MLP also provides a significant speedup for the Jacobian computation wrt global pose: approximate Jacobian ($12\times$) and canonical Jacobian ($860\times$).

\section{Method}
\label{sec:method}
In this section, we describe LUTI-MLP, a novel point feature embedding architecture that significantly speeds up the embedding and its Jacobian computation at test time.
After preliminaries (\Sec\ref{subsec:preliminary}), we first review PointNet \cite{qi2017pointnet} (\Sec\ref{subsec:pointnet}), which enabled the permutation invariant modeling of the points set, and then formalize the problems we consider (\Sec\ref{subsec:problem}).
Next, we introduce LUTI-MLP, a specifically designed tabulated MLP (\Sec\ref{subsec:luti_mlp}), which enables the efficient linear computation of the nonlinear embedding at test time (\Sec\ref{subsec:embedding}), and it also realize highly efficient Jacobian computation of the embedding function wrt global pose in $\mathfrak{se}(3)$ at test time (\Sec\ref{subsec:jacobian}).
Finally, training method for LUTI-MLP are discussed in (\Sec\ref{subsec:training}).

\subsection{Preliminaries}
\label{subsec:preliminary}
We consider a 3D geometric point set, $\mathbf{P}=\{\mathbf{p}_i |i = 1, ..., N\}$, where each point $\mathbf{p}_i$ is a vector of its $(x, y, z)$ coordinate.
We want to model the following relation between the input point set and its output as
\begin{equation}
\mathbf{o}=f(\mathbf{P}),
\end{equation}
where output $\mathbf{o}$ could be the class label for the 3D object classification task, or it could be the point-wise label for the semantic segmentation task.
To realize such functions using neural networks, a permutation of input points would be a huge problem.
A simple MLP or RNN \cite{Rumelhart:1986we} trained with randomly permuted sequences cannot scale to thousands or tens of thousands of input elements \cite{qi2017pointnet}. 
Voxelization, followed by a 3D convolutional neural network (3D-CNN) \cite{tran2015learning,maturana2015voxnet,roynard2018classification}, is a commonly used strategy for processing a 3D point set, but its computational complexity prevents it from scaling to the higher resolution, and it may suffer from performance degeneration induced by information loss from the voxelization. 

\subsection{Review of PointNet \cite{qi2017pointnet} and Variants}
\label{subsec:pointnet}
The basic idea of PointNet (\Fig\ref{fig:overview}, top) is to embed each input-point feature independently into high-dimensional feature space as
\begin{equation}
\label{eq:mlp_basic}
\mathbf{z}_i=\phi(\mathbf{p}_i),
\end{equation}
where  $\phi:  \mathbb{R}^3 \rightarrow \mathbb{R}^K$ is a nonlinear embedding function realized with MLP.
The set of embedded feature $\mathbf{Z}:\{\mathbf{z}_1,...,\mathbf{z}_N\}$ is then aggregated to global feature $\mathbf{a}$ by the symmetric function $\max$ as
$\mathbf{a}=\max\{\mathbf{z}_1,...,\mathbf{z}_N\}$,
where  $\max:  \mathbb{R}^K\times ...\times  \mathbb{R}^K \rightarrow \mathbb{R}^K$. 
Then, the global feature $\mathbf{a}$ is processed by another function $g$ to compute the output as $\mathbf{o}=g(\mathbf{a})$ which may differ according to the type of task.
Because of the symmetry of the aggregation function, the permutation of input point clouds does not change the global feature $\mathbf{a}$, so output $\mathbf{o}$ does not change as a result.

\prg{Object Classification}
Global feature is fed directly to the classification MLP
(mlp-c in \Fig \ref{fig:overview}, top) to output the scores for the candidate classes \cite{qi2017pointnet,qi2017pointnetplusplus,qi2018frustum,zhou2018voxelnet,lang2019pointpillars}.

\prg{Point-wise Classification}
The concatenation of point-wise embedded features $\mathbf{Z}^{feat}$ and global feature $\mathbf{a}$ is processed by another MLP to output point-wise score \cite{qi2017pointnet,qi2017pointnetplusplus}.

\prg{Point Set Registration}
The proximity of global features $|\mathbf{a}_\mathcal{S}-\mathbf{a}_\mathcal{T}|$, where  $\mathbf{a}_\mathcal{S}$ and $\mathbf{a}_\mathcal{T}$ are computed from source  and target point set,  relates to the geometric proximity of the set in input space. It is used to compute the geometric transformation between the sets \cite{aoki2019pointnetlk}. 

\subsection{Problem Statement}
\label{subsec:problem}
Our research goal is to realize efficient computation of the nonlinear embedding $\mathbf{z}\in \mathbb{R}^K$ and its differential operation wrt input point coordinates $\mathbb{R}^3$ and global pose coordinates $\mathfrak{se}(3)$.
When we consider processing 3D point clouds for real-time recognition, the embedding of thousands or tens of thousands of inputs must be computed dozens of times within a second at a specific frame rate (e.g., $100$ Hz).
This means the embedding function $\phi$ realized by MLP must be evaluated millions of times per second.
It is quite difficult because MLP consists of layers of matrix-vector product operation, which requires millions of dot products per a point input.
Thus, the speed of the embedding is critical.
Similarity,  Jacobian of global feature $\mathbf{a}$ wrt geometric transformation in $\mathfrak{se}(3)$ needs to be computed efficiently in point-set registration problems \cite{aoki2019pointnetlk}.

\subsection{LUTI-MLP}
\label{subsec:luti_mlp}
We consider replacing MLP with a LUT to speed up the embedding by avoiding the matrix-vector product operation. 
LUT  operation could be more than two orders of magnitude faster than MLP.
If the resolution of the input point is discrete, it is actually possible to replace MLP with a LUT by pre-computing the relation $\mathbf{z}=\phi(\mathbf{p})$ for every possible input points $\mathbf{p}$.
However, memory footprint prevents a realizing of this approach. 
For example, $1024\times  1024\times 1024$ sized LUT requires $4$ Terabytes of memory\footnote{See supplementary (G) for more detail.}  when $K=1024$, which is infeasible even with the latest high-end GPUs.
Simply reducing the input resolution ruins the vital characteristic of point feature embedding of \eqref{eq:mlp_basic} that can model continuous input, and it degrades the performance as in the case for low-resolution voxel representation for 3D-CNN.

Instead of duplicating or approximating a trained MLP of vanilla PointNet using LUT, we propose a novel end-to-end (E2E) trainable neural network framework; we call this as LUTI-MLP, that uses tabulated MLP to transform point-coordinate inputs into high-dimensional features.
LUTI-MLP computes the embedding as a weighted sum of the basis function $\phi_D(\mathbf{\bar{p}}^j)$ which is output from MLP evaluated with an input on lattice $\mathbb{D}^3$, where the weight is computed using the spatial proximity between input $\mathbf{p}_i$ and discrete grid $\mathbf{\bar{p}}^j_i$  on the lattice: 
\begin{equation}
\label{eq:lutimlp_basic}
\mathbf{z}_i=\phi(\mathbf{p}_i)=\sum_j w^{(j)}(\mathbf{p}_i,\mathbf{\bar{p}}^{(j)}) \phi_{D}(\mathbf{\bar{p}}^{(j)}).
\end{equation}
Notice that the basis functions $\phi_D(\bar{\mathbf{p}}^{(j)})$ are locally independent from input coordinate $\mathbf{p}_i$ in contrast to the direct dependance  on $\phi$ in \eqref{eq:mlp_basic}.
At test time, this formulation leads to the drastic speedup of embedding and Jacobian computation by using pre-computed basis functions $\phi_D(\bar{\mathbf{p}}^{(j)})$ rather than evaluating MLP.  
Note that the embedding function $\phi(\mathbf{p})$ and its output $\mathbf{z}$ of LUTI-MLP in \eqref{eq:lutimlp_basic}  are different from \eqref{eq:mlp_basic}  that is realized by single MLP defined over continuous Euclidean space $\mathbb{R}^3$.
The discretization $D$ and weight function $w$ control the expressiveness of the function.
We experimentally confirmed that even a very coarse discretization ($D=4$) in combination with simple trilinear interpolation as $w$ yields excellent performance comparable to that of PointNet using MLP, which consists from layers of matrix-vector product (\Sec\ref{sec:exp}).

\subsection{Fast Embedding Computation at Testing}
\label{subsec:embedding}
At test time, the embedding feature $\mathbf{z}$ in \eqref{eq:lutimlp_basic} is computed as a linear combination of pre-computed basis function  $\phi_D(\bar{\mathbf{p}}^{(j)})$ stored on LUT, the look-up followed by interpolation are much more cheaper than evaluating MLP that involves computationally intensive matrix-vector product operation.
We choose trilinear interpolation as a weight function $w$ for input in  $\mathbb{R}^3$.
The network architecture of PointNet for classification with our LUTI-MLP at test time is illustrated in \Fig\ref{fig:overview}, middle.
The continuous input point $\mathbf{p}_i$ is discretized to 8  neighborhood points, $\bar{\mathbf{p}}_i^{(000)},...,\bar{\mathbf{p}}_i^{(111)}$on the lattice, 
and 8 pre-computed basis function $\phi_D(\bar{\mathbf{p}}^{(j)}_i)$ corresponds to these discretized 8 input points are looked-up as  $[\bar{\mathbf{z}}_i^{(000)},...,\bar{\mathbf{z}}_i^{(111)}]=[\phi_D(\bar{\mathbf{p}}_i^{(000)}),...,\phi_D(\bar{\mathbf{p}}_i^{(111)})]$.
Then, these 8 basis functions are trilinearly interpolated using the Euclidean distance between the input and the neighborhood points on the lattice as the weight, that is,
\begin{multline}
\label{eq:trilinear}
\mathbf{z}_i
=\bar{\mathbf{z}}_i^{(000)}\bar{d_x}\bar{d_y}\bar{d_z}
+\bar{\mathbf{z}}_i^{(001)}\bar{d_x}\bar{d_y}\aar{d_z}\\
+\bar{\mathbf{z}}_i^{(010)}\bar{d_x}\aar{d_y}\bar{d_z}
+\bar{\mathbf{z}}_i^{(011)}\bar{d_x}\aar{d_y}\aar{d_z}\\
+\bar{\mathbf{z}}_i^{(100)}\aar{d_x}\bar{d_y}\bar{d_z}
+\bar{\mathbf{z}}_i^{(101)}\aar{d_x}\bar{d_y}\aar{d_z}\\
+\bar{\mathbf{z}}_i^{(110)}\aar{d_x}\aar{d_y}\bar{d_z}
+\bar{\mathbf{z}}_i^{(111)}\aar{d_x}\aar{d_y}\aar{d_z},
\end{multline}
where $d=p \rceil-p$ and $\bar{d}=p-p\rfloor$ are the Euclidean distance to the neighbors on the lattice.

\paragraph{Feature Transformation}
To get some kind invariance against a geometric transformation of input, the authors of PointNet \cite{qi2017pointnet} also proposed the feature transformation structure to canonicalize the input point set.
They proposed to integrate another PointNet structure (T-Net) to estimate the affine transformation matrix inspired by STN \cite{jaderberg2015spatial}, and apply this transformation directly to the coordinates of the input points. 
They used two T-Net for different layers, one for the input layer as input transform (IT) and the other for the middle of MLP as feature transform (FT).
In this paper, we mainly focus on the network with and without IT, and do not deal with the one with FT for two reasons: 1) simplicity of training\footnote{FT needs orthogonal regularization for the affine parameter to get a better performance; otherwise, it deteriorated the performance instead.}, and 2) in our preliminary experiments with IT and FT (with orthogonality regularization), we saw no performance gain\footnote{The classification accuracy of ShapeNet40 is 85.87\%, and this is slightly worse than vanilla PointNet with IT.}.
The embedding MLP of T-Net can be replaced by our LUTI-MLP to speed up the computation of the transformation matrix.
We modified the original T-Net (IT) slightly\footnote{Detailed network architecture for T-Net (IT) and T-Net (FT), and additional results from using T-Net (FT) are shown in supplementary (A).} by inserting $\tanh$ at the end of T-Net so that the input to the LUTI-MLP on the following layer is also bounded within unit cube.

\subsection{Fast Jacobian Computation wrt Pose at Testing}
\label{subsec:jacobian}
Using the formulation of LUTI-MLP in \eqref{eq:lutimlp_basic}, Jacobian of each element of embedded feature $\mathbf{z}$ wrt the input coordinates  $(x,y,z) \in \mathbb{R}^3$ in Euclidean space is computed quite efficiently.
In the case of trilinear interpolation, the Jacobian is derived as the derivative of \eqref{eq:trilinear}:

\begin{multline}
\label{eq:jacobian}
\frac{\partial \phi(\mathbf{p})_i}{\partial x}
=(\bar{\mathbf{z}}_i^{(000)}-\bar{\mathbf{z}}_i^{(100)})(\bar{d_y}\bar{d_z})
+(\bar{\mathbf{z}}_i^{(010)}-\bar{\mathbf{z}}_i^{(110)})(\aar{d_y}\bar{d_z})\\
+(\bar{\mathbf{z}}_i^{(001)}-\bar{\mathbf{z}}_i^{(101)})(\bar{d_y}\aar{d_z})
+(\bar{\mathbf{z}}_i^{(011)}-\bar{\mathbf{z}}_i^{(111)})(\aar{d_y}\aar{d_z}),
\end{multline}
which is a bilinear interpolation of the difference of the  basis functions $\phi_D(\bar{\mathbf{p}}^{(j)})$ that is also pre-computable.
Jacobian wrt $y$ and $z$ are computed similarly.
This Jacobian wrt $\mathbb{R}^3$ is easily pulled-back to Jacobian wrt geometric coordinates $\boldsymbol{\xi}$  in  $\mathfrak{se}(3)$:
\begin{align}
\label{eq:dpdxi}
\frac{\partial \mathbf{p}}{\partial\boldsymbol{\xi}}=
\left( {\begin{array}{*{20}{c}}
  0&{ - z}&{ y} \\ 
  { z}&0&{-x} \\ 
  {-y}&x&0 
\end{array}\begin{array}{*{20}{c}}
  1&0&0 \\ 
  0&1&0 \\ 
  0&0&1
\end{array}} \right).
\end{align}
Using the relation of \eqref{eq:dpdxi},  the analytical Jacobian of embedded feature $\mathbf{z}$ wrt. $\boldsymbol{\xi}$ is given as
\begin{equation}
\label{eq:dadxi}
\frac{\partial \phi(\mathbf{p})}{\partial \boldsymbol{\xi}}=\frac{\partial  \phi(\mathbf{p})}{\partial \mathbf{p}}\frac{\partial \mathbf{p}}{\partial \boldsymbol{\xi}}.
\end{equation}

\paragraph{Application to Point Set Registration}
PointNetLK \cite{aoki2019pointnetlk} proposed to use global feature for 3D point cloud registration.
It uses the global feature from two point sets (source $\mathbf{P}_{\mathcal{S}}$ and target  $\mathbf{P}_{\mathcal{T}}$)  to measure the proximity of the pair as
\begin{equation}
\label{eq:invj}
\mathbf{r}=[\max(\phi(\mathbf{G\cdot P}_{\mathcal{S}})-\max(\phi(\mathbf{P}_{\mathcal{T}}))].
\end{equation}
The geometric transformation, $\mathbf{G}\in SE(3)$ between sets is estimated as a minimizer of the residual  $\mathbf{r}$ wrt $\mathbf{G}$. 
Jacobian $\mathbf{J}$ for the aggregated embedded feature $\max(\phi(\mathbf{P}_{\mathcal{T}}))$ wrt $\boldsymbol{\xi}\in\mathfrak{se}(3)$ is used in inverse compositional formulation \cite{baker2004lucas} as
$\boldsymbol{\Delta\xi}=\mathbf{J}^{\dagger}\mathbf{r},$
where $\mathbf{J}^{\dagger}$ is Moore-Penrose inverse of  $\mathbf{J}$ which is computed only once  for given target point set  $\mathbf{P}_{\mathcal{T}}$, and the residual $\mathbf{r}$ is computed iteratively using $\mathbf{G}$  until convergence, where it is updated as $\Delta \mathbf{G}=\exp \left(\sum_{k=1,...,6} \Delta\boldsymbol{\xi}_{k} \mathbf{T}_{k}\right)$.
The acceleration of LUTI-MLP can be used to compute $\phi(\mathbf{G\cdot P}_{\mathcal{S}})$ in the iterative residual computation, and it can also be used for Jacobin computation (in two ways), which are discussed below.

\prg{Approximate Jacobian}
Because of the difficulties for computing the analytical Jacobian of MLP, the authors in \cite{aoki2019pointnetlk} used the finite difference approximation as
\begin{equation}
\label{eq:approx_lk}
\mathbf{J}_{k}=\frac{\max(\phi(\exp (-t_{k} \mathbf{T}_{k})\cdot \mathbf{P}_{\mathcal{T}}))-\max(\phi(\mathbf{P}_{\mathcal{T}}))}{t_{k}},
\end{equation}
where $t_k$ is the small perturbation in the $k$-th element of the $\mathfrak{se}(3)$ parameter.
We can replace embeddings $\phi$ realized as MLP with our LUTI-MLP to speed up the computation of this finite-difference.

\prg{Canonical Jacobian}
Instead of the numerical differentiation, \eqref{eq:dadxi} gives the efficient analytical Jacobian of global feature $\mathbf{a}$ wrt  $\boldsymbol{\xi}$ as
\begin{align}
\label{eq:canonic_lk}
    \mathbf{J}&=\frac{\partial \mathbf{p}}{\partial \boldsymbol{\xi}}\frac{\partial}{\partial \mathbf{p}}[\max(\phi(\mathbf{G}^{-1} \cdot \mathbf{P}_{\mathcal{T}}))].
\end{align}
Contrary, computing the analytical Jacobian of MLP wrt $\mathbf{p}$ and $\boldsymbol{\xi}$  is quite complicated, requiring the partial derivative computation of each element of embedding feature $\mathbf{z}$ wrt nonlinear MLP.
It is computed by looping the backpropagation from each element of $\mathbf{z}$  for $K$ times.

\subsection{Training LUTI-MLP}
\label{subsec:training}
In \Fig\ref{fig:overview} on the bottom, network architecture of  LUTI-MLP at training are shown.
LUTI-MLP is differentiable and can be trained E2E because $\phi_D$ and the interpolation of \eqref{eq:lutimlp_basic} are differentiable. 
The basis function $\phi_D(\bar{\mathbf{p}}^{(j)})$, which is actually a MLP, is trained using standard back-propagation using the error from upper layers.
The back-propagated error on the interpolated embedding $\mathbf{z}_i$  are distributed to the basis function $\phi_D(\bar{\mathbf{p}}^{(j)}_i)$  on lattice, according to the weight $w^{(j)}(\mathbf{p}_i,\mathbf{\bar{p}}^{(j)}) $ which is used in the forward computation.
Because the same interpolation is used at training and test time, the embedding $\mathbf{z}$ computed using pre-computed $\phi_D$ on  LUT at test time is exactly the same as the one computed using MLP $\phi_D$ at training.
As discussed in \Sec\ref{sec:exp}, the simple application of trilinear interpolation of \eqref{eq:trilinear} to approximate the trained PointNet with MLP yields poor performance.
We experimentally revealed that E2E training is key to achieve an excellent performance comparable to that of PointNet with MLP.

\section{Experiments}
\label{sec:exp}
The experiments are divided into four parts. 
First, we see the speedup gain for embedding and Jacobian computation by applying LUTI(-MLP) to the point-set embedding function (\Sec \ref{subsec:exp_speed}). 
Second, we show the applicability of our LUTI-MLP to several architectures, such as ones designed for object classification, point-wise classification, and point-cloud registration (\Sec \ref{subsec:exp_app}).
Then, we provide an intensive architectural analysis to see how the discretization ratio or architectural design affects the performance (\Sec \ref{subsec:analysis}). 
Last, we visualize what the embedding network with different discretization learns (\Sec \ref{subsec:exp_vis}).

\prg{Implementation}
For all the experiments, we used PyTorch \cite{paszke2017automatic} library for training and testing.
We  followed the original description for the basic network architecture and training procedure of PointNet \cite{qi2017pointnet}
for classification and segmentation, and PointNetLK \cite{aoki2019pointnetlk} for point set registration.
The only difference from the original description is the use of  LUTI-MLP instead of MLP for embedding, and a slight modification of T-Net (IT) (\Sec\ref{subsec:embedding}). 
No other modification, such as hyper-parameter tuning, was performed.

\subsection{Computational Speed Analysis of LUTI}
\label{subsec:exp_speed}
We evaluated the speedup gain by using our LUTI(-MLP) instead of MLP for embedding.
For this speed benchmarking, we implemented the Nvidia CUDA kernel for trilinear interpolation (embedding)  and bilinear interpolation (Jacobian).
The benchmark was done with and without cudnn acceleration for MLP.
Batch normalization \cite{ioffe2015batch} used in MLP was integrated into the matrix-vector products layer for a fair comparison. 
For both experiments, we report the average wall-clock time to process randomly generated $1000$ points in the unit cube. 
We used Nvidia V100 GPU for all speed benchmarking.

\paragraph{Embedding}
\begin{table}
\caption{\label{table:speed}
\textbf{Time complexity of feature embedding.}
We compare the computational time ($\mu$s) of LUTI and MLP as the embedding function of PointNet.
The benchmark was done with and without cudnn acceleration for MLP.
Our LUTI version of embedding is much faster ($80\times$) than MLP with cudnn.
Despite the number of lookup and linear interpolation of LUTI are the same across different $D$, its computation time increase as $D$ increase for small $D$ and became almost constant for $D>32$.
It is because, the memory access tend to take more time for larger LUT.
}

\begin{center}
\begin{adjustbox}{width=1\columnwidth}
\begin{tabular}{cc||cccccc}
& & \multicolumn{6}{c}{ Number of discretization $D$} \\
\hline
Alg. & cudnn & inf & 64 & 32 & 16 & 8 & 4 \\
\hline
MLP & \checkmark & 4330 & - & - & - & - & - \\
MLP & & 4350 & - & - & - & - & - \\ \hdashline
LUTI & - & - & 71.8 & 71.6 & 69.8 & 53.6 & 46.8 \\
\end{tabular}
\end{adjustbox}
\end{center}
\end{table}

\Tab\ref{table:speed}  compares the computational complexity of embedding between the MLP and our LUTI-MLP at test time.
The speedup gain from LUTI acceleration is more than $80\times$ compared with the highly optimized cudnn implementation of MLP.
Please note that MLP version slows down as the number of layers or intermediate channel increases, while LUTI is invariant to the architecture.

\paragraph{Jacobian}
\begin{table}
\caption{\label{table:speed_jac}
\textbf{Time complexity of  Jacobian computation and pose update.}
We compare the computational time of four different types of Jacobians and corresponding pose estimation.
From top to bottom,  the finite-difference  Jacobian of MLP \cite{aoki2019pointnetlk} and LUTI-MLP, and canonical  Jacobian of MLP and  LUTI-MLP.
Our LUTI accelerates the computation of approximate Jacobian ($12\times$), canonical Jacobian ($860\times$), pose estimation using approximate Jacobian ($37\times$), and pose estimation using canonical Jacobian ($200\times$) respectively compared with MLP with cudnn. 
$D=8$ is used for LUTI.
}
\begin{center}
\begin{tabular}{ccc||cc}
\multicolumn{2}{c}{ Alg. } & cudnn & Jac. ($\mu$s) & All (ms) \\
\hline
Approx. & MLP & \checkmark & 4860 & 91.46 \\
Approx. & MLP & & 5340 & 92.34 \\ \hdashline
Approx. & LUTI & - & $\mathbf{399}$ & $\mathbf{2.47}$ \\
\hline
Canonic. & MLP & \checkmark & 437000 & 523.6 \\
Canonic. & MLP & & 786000 & 873 \\ \hdashline
Canonic. & LUTI & - &  $\mathbf{503}$ &  $\mathbf{2.58}$\\
\end{tabular}
\end{center}
\end{table}
 
\Tab\ref{table:speed_jac}  compares the computational complexity of Jacobian computation between the MLP and our LUTI-MLP at test time.
Besides, the computational speed of the entire pose estimation, including $20$ iterations of the residual evaluation of \eqref{eq:invj}, are also reported.
We compared four different types of  Jacobian computation,  the finite-difference  Jacobian of MLP \cite{aoki2019pointnetlk} and LUTI-MLP, and canonical  Jacobian of MLP and  LUTI-MLP.
The computation of the approximate Jacobian includes the first term of the numerator of \eqref{eq:approx_lk}, $\max(\phi\left(\exp \left(-t_{k} \mathbf{T}_{k}\right) \cdot \mathbf{P}_{\mathcal{T}}\right)$, which is warping of point clouds, followed by embedding and aggregation by $\max$.
The computation of canonical Jacobian using LUTI includes bilinear interpolation of \eqref{eq:jacobian} and the index copy operation using $\argmax$, which is assumed to be pre-computed.
The computation of the canonical Jacobian using MLP requires $K$ times of backpropagation\footnote{For this benchmarking, we use the parallelization technique disclosed in https://gist.github.com/sbarratt (torch\_jacobian.py) to accelerate the multiple backward computations.}, and the same index operation is followed. 
Our LUTI accelerates the computation of the approximate Jacobian ($12\times$), canonical Jacobian ($860\times$), pose estimation using approximate Jacobian ($37\times$), and pose estimation using canonical Jacobian ($200\times$) compared with MLP with cudnn. 

\subsection{Applications}
\label{subsec:exp_app}
We demonstrate how our LUTI can be integrated in several architectures, such as 3D object classification, object part segmentation, and point-set registration. 
We evaluated our PointNet and PointNetLK with LUTI-MLP against the original network using MLP for embedding.
In summary, we observe the orders of magnitude speedup without performance deterioration for different types of architectures and applications.
We use a discretization of $D=8$ for all applications.

\paragraph{3D Object Classification}
Given a 3D point cloud, the task is to assign an object class label for the point set. 
We evaluated our model on the ModelNet40 \cite{wu20153d} shape classification benchmark. 
There are $12,311$ CAD models from 40 object categories, split into $9,843$ for training and $2,468$ for testing.
We randomly sampled $1,000$ points and normalized them into a unit cube. 
During training, we followed the same data argumentation protocol of the PointNet \cite{qi2017pointnet},  augment the point cloud on-the-fly by randomly rotating the object along the up axis, and jitter the position of each point by Gaussian noise with zero mean and 0.02 standard deviation.
The results are summarized in \Fig \ref{fig:effect_D}.
Our PointNet using LUTI-MLP (\textit{LUTI-MLP E2E})  slightly outperformed the original network (\textit{MLP}) while achieving more than $80\times$ speedup for embedding.
Detailed architectural analysis using the same task is discussed in \Sec\ref{subsec:analysis}.

\paragraph{3D Object Part Segmentation}
\begin{table}
\caption{\label{table:shapenet_digest}
$\textbf{Segmentation results on ShapeNet part dataset.}$ 
Our PointNet with LUT-MLP achieves comparable results as original PointNet with MLP while achieving significant speed-up.
Metric is mIoU(\%) on points. 
}

\begin{center}
\begin{adjustbox}{width=1\columnwidth}
\begin{tabular}{cc|c|cccc}
& IT  & mean & chair & table & aero & lamp \\
\# shapes & & & 3758 & 5271 & 2690 & 1547 \\\hline
MLP &  & 84.73 & 88.09 & 92.59 & 79.80 & 78.44 \\
MLP & \checkmark  & 84.38 & 88.09 & 91.32 & 76.19 & 81.90 \\ \hdashline
LUTI &  & 85.56 & 88.58 & 92.54 & 79.28 & 81.83 \\
LUTI & \checkmark  & 85.36 & 87.89 & 91.52 & 82.01 & 80.03 \\
\end{tabular}
\end{adjustbox}
\end{center}
\end{table}
Given a 3D point cloud, the task is to assign a part category label to each point. 
We evaluate on the ShapeNet part data set from \cite{shapenet2015}, which contains $16,881$ shapes from $16$ categories.
Out of the $16$ categories,  we trained and evaluated for \textit{Chair}, \textit{Table}, \textit{Aeroplane}, and  \textit{Lamp} categories which have the largest number of models (more than $1500$).
Local features $\mathbf{Z}^{feat}$ for LUTI is stored in the same table with a global feature; thus, the table size for this experiments is $D\times D\times D\times (1024+64)$.
The results are summarized in \Tab\ref{table:shapenet_digest}.
Our network with LUTI-MLP showed comparable performance as the original PointNet while achieving more than $80\times$ speedup for embedding\footnote{We report results from different discretization on supplementary (B).}.

\paragraph{3D Point Cloud Registration}
\label{subsubsec:PTLK}
Given source ($\mathbf{P}_{\mathcal{S}}$) and target  ($\mathbf{P}_{\mathcal{T}}$) 3D point cloud, the task is to estimate the 3D geometric transformation in $SE(3)$ between the sets.
We followed the same point set registration algorithm of PointNetKL \cite{aoki2019pointnetlk}.
As discussed in \Sec\ref{sec:method}, LUTI is used to accelerate iterative residual computation \eqref{eq:invj} and for Jacobian computation either approximately \eqref{eq:approx_lk} or analytically \eqref{eq:canonic_lk}.
We compared the registration accuracy of PointNetLK using our LUTI-MLP and PointNetLK using MLP for embedding.
Both variants using LUTI-MLP (one use approximate and the other use canonical Jacobian) achieved accuracy comparable to that of the original PointNetLK using MLP\footnote{The detailed evaluation results are shown in supplementary (C)} while achieving orders of magnitude speedup (\Tab\ref{table:speed_jac}). 
In addition to the efficiency, the canonical Jacobian has practical advantages that it has no hyperparameter $t$ in \eqref{eq:approx_lk} (a small perturbation) for computing the finite difference whose appropriate value may differ depending on the network architecture, or data to process. 

\subsection{Architecture Design Analysis}
\label{subsec:analysis}
\begin{figure}[!h]
\includegraphics[width=1\columnwidth]{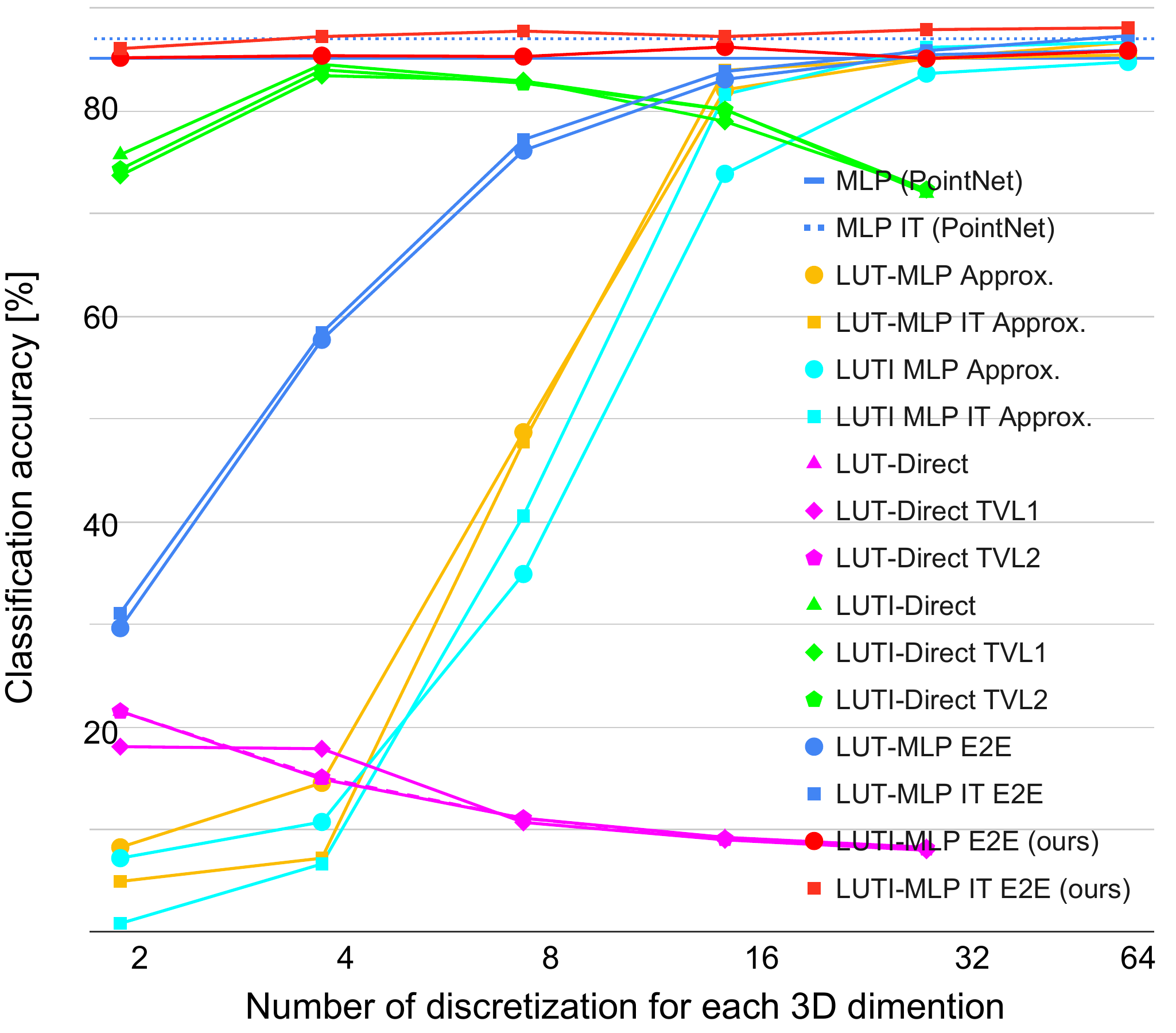}
\caption{
\label{fig:effect_D}
\textbf{Effects of table size and architectural design for the object classification performance on the Model-Net40 test set.}
Lines without marker represents the results from baseline PointNet without and with T-Net for IT.
The metric is overall classification accuracy.
Our PointNet with LUTI-MLP shows comparable or slightly better results against naive PointNet across all range of $D$ while achieving significant speedup.
See the main text for the explanation of each algorithm.
}
\end{figure}

We used the same object classification task discussed in \Sec\ref{subsec:exp_app} to analyze the effect of architectural difference and discretization degree.
We compared the following $7$ different architectures with different discretization.
For all architectures except \textit{LUT-Direct} and \textit{LUTI-Direct}, we experimented with variants with and without T-Net (IT).
The results are summarized in \Fig \ref{fig:effect_D}.
All models were trained for a total of  $200$ epochs using the same scheduling as described in the original paper \cite{qi2017pointnet}.

\prg{\textit{MLP}}
It is a re-implementation of the PointNet with a slight modification on T-Net (IT), and we consider it a baseline architecture.
We get similar results as in the original paper.

\prg{\textit{LUT-MLP Approx.} / \textit{LUTI-MLP Approx.}}
These variants approximate trained MLP of vanilla PointNet using LUT or LUTI.
\textit{LUT-MLP Approx.} simply discretized input points to $\mathbb{D}^3$.
\textit{LUTI-MLP Approx.} used the interpolation of \eqref{eq:trilinear}, but no E2E training for the basis function was conducted.
As expected, the performance of both architectures dropped drastically as the discretization became coarser (around $D=16$),  because the distance between \textit{true} embedding $\phi (\mathbf{p})$  and the approximated embedding either using nearest neighbor lookup or interpolation of \eqref{eq:trilinear} became large.   
Contrary to our expectations, \textit{LUTI-MLP Approx.} with interpolation performed slightly worse than \textit{LUT-MLP Approx.} without interpolation.
This result may partly be attributed to the fact that the distance between the  \textit{true}  embedding $\phi (\mathbf{p})$ and approximated embedding from LUTI can be larger than that of the LUT (nearest-neighbor lookup), where the linear assumption was not satisfied.

\prg{\textit{LUT-Direct}  / \textit{LUTI-Direct}\footnote{These two variants were not evaluated for $D=64$, because the huge memory footprint for training cannot fit into our GPU.}}
These variants does not incorporate PointNet like MLPs as a proxy for training the basis function on LUT, but directly train $D\times D\times D\times K$ table as a training parameter $\mathbf{W}$.
\textit{LUTI-Direct} is equivalent to training the network of LUTI-MLP at test time (\Fig \ref{fig:overview}, middle).
The table was updated directly using the backpropagated classification error from mlp-c.
In this direct training, elements of the table $\mathbf{W}$, which had no corresponding input, receives no gradient signal for training, which may negatively affect the generalization to unseen point coordinates.
Therefore, we examined the use of  regularization using the total variation (TV)  with $p$ norm for $\mathbf{W}$ (TVL1 for $p=1$, TVL2 for $p=2$). 
The TV was evaluated using  3D spatial neighbor $\mathcal{N}$  as
$
T V(\mathbf{W})=\sum_{i, k \in \mathcal{N}}\left\|\mathbf{w}_{i}-\mathbf{w}_{j}\right\|^{p}.
$
The results from \textit{LUT-Direct} were overwhelmingly worse than those for the other variants especially when we used fine discretization.
\textit{LUT-Direct} updates only looked-up parameters. 
Therefore, it is difficult to train when the spatial resolution of the table is fine because the gradients become sparse. 
Even if the parameters receive the gradients wrt the regularization, the gradients from the regularizer do not improve classification accuracy directly.
Surprisingly, however,  \textit{LUTI-Direct}, which incorporates LUTI, performed reasonably well for coarse discretization.
It may because the interpolation of LUTI helps the error signal to flow into the neighboring element with an appropriate weight according to the proximity of the input to the neighbor lattice, and the coarse table structure works as regularizer.

\prg{\textit{LUT-MLP E2E}}
This variant use discretized input points on $\mathbb{D}^3$ as in \textit{LUT-MLP Approx.}, but trained E2E.
The discretization loses information on input similar to the case of CNN-based architecture \cite{tran2015learning} with coarse voxelization performs poorly.     
As expected, the performance dropped as the discretization became coarser (around $D=16$).

\prg{\textit{LUTI-MLP E2E}}
This variant is our proposed architecture, which incorporates our LUTI-MLP and trains the network E2E.
For this variant, we used a pre-trained parameter from the \textit{MLP} at $100$ epochs and then trained for another $100$ epochs\footnote{The primary purpose of pre-training was to see the evolution of the embedding feature in \Sec \ref{subsec:exp_vis}. In practice, this pre-training has little impact on classification accuracy, as shown in the supplementary (D). }.
This variant performed slightly better than the original architecture for a wide range of $D$ and outperformed other variants by a large margin.
From another perspective, the comparison with \textit{LUTI-Direct} suggests that the use of MLP coupled with LUT itself provides a suitable training method for training LUT for interpolation of point-feature embedding.

\subsection{Visualizing the PointNet Embedded Space}
\label{subsec:exp_vis}
\begin{figure}[!h]
\includegraphics[width=1\columnwidth]{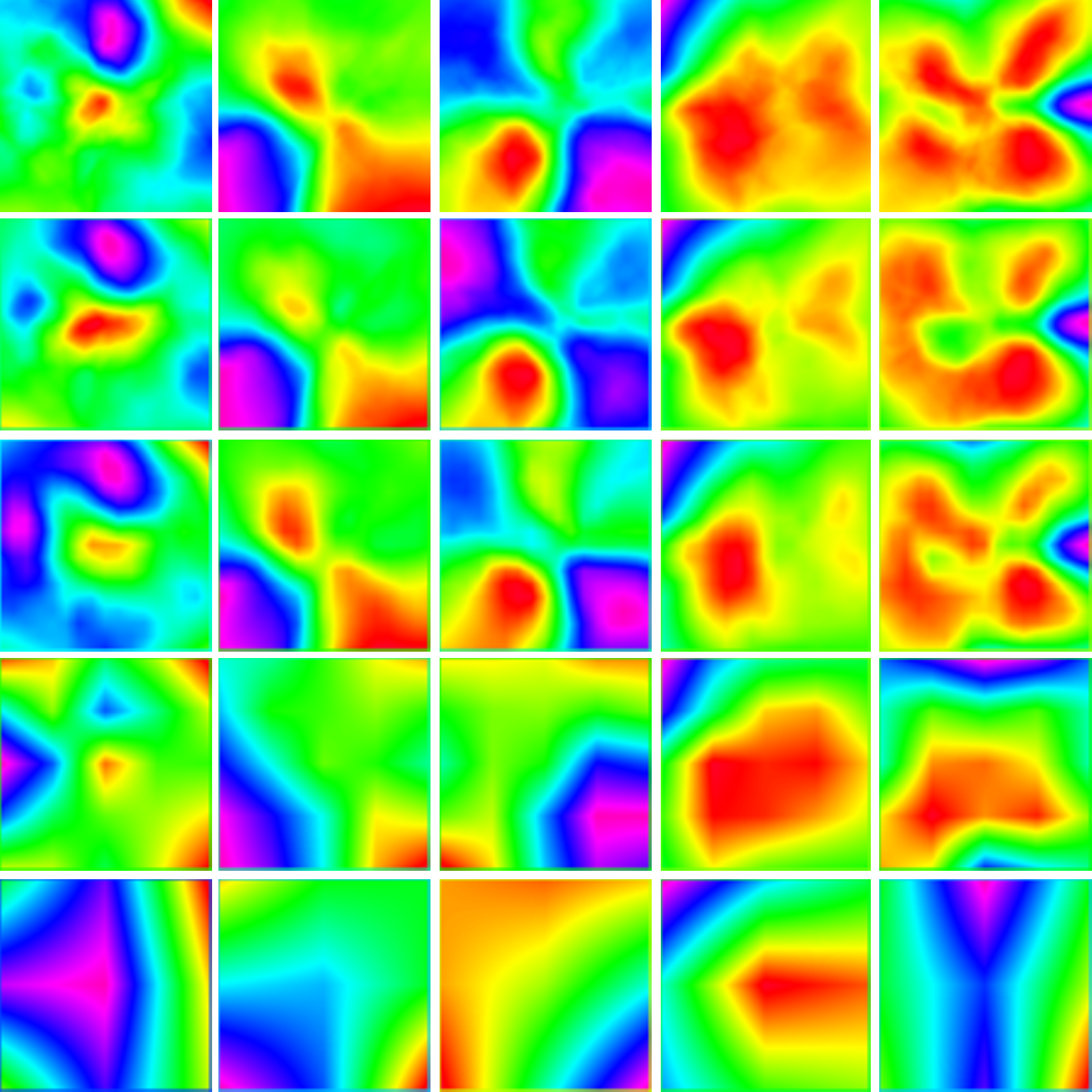}
\caption{
\label{fig:vis_emb_mlp}
\textbf{Visualization of the trained embedding feature of PointNet using MLP (\textit{MLP}) and PointNet with LUTI-MLP (\textit{LUTI-MLP E2E}).}
The results are reported for different discretizations $D$ for \textit{LUTI-MLP}.
From top to bottom, \textit{MLP} (PointNet), \textit{LUTI-MLP} with $D=64, 16, 4$ and $2$.
Note that the  \textit{LUTI-MLP} use pre-trained MLP from naive PointNet (\textit{MLP} on \Sec \ref{subsec:analysis}).
Five randomly selected channels of the slice on the $z=0$ plane are shown.
}
\end{figure}
\begin{figure}[!h]
\includegraphics[width=1\columnwidth]{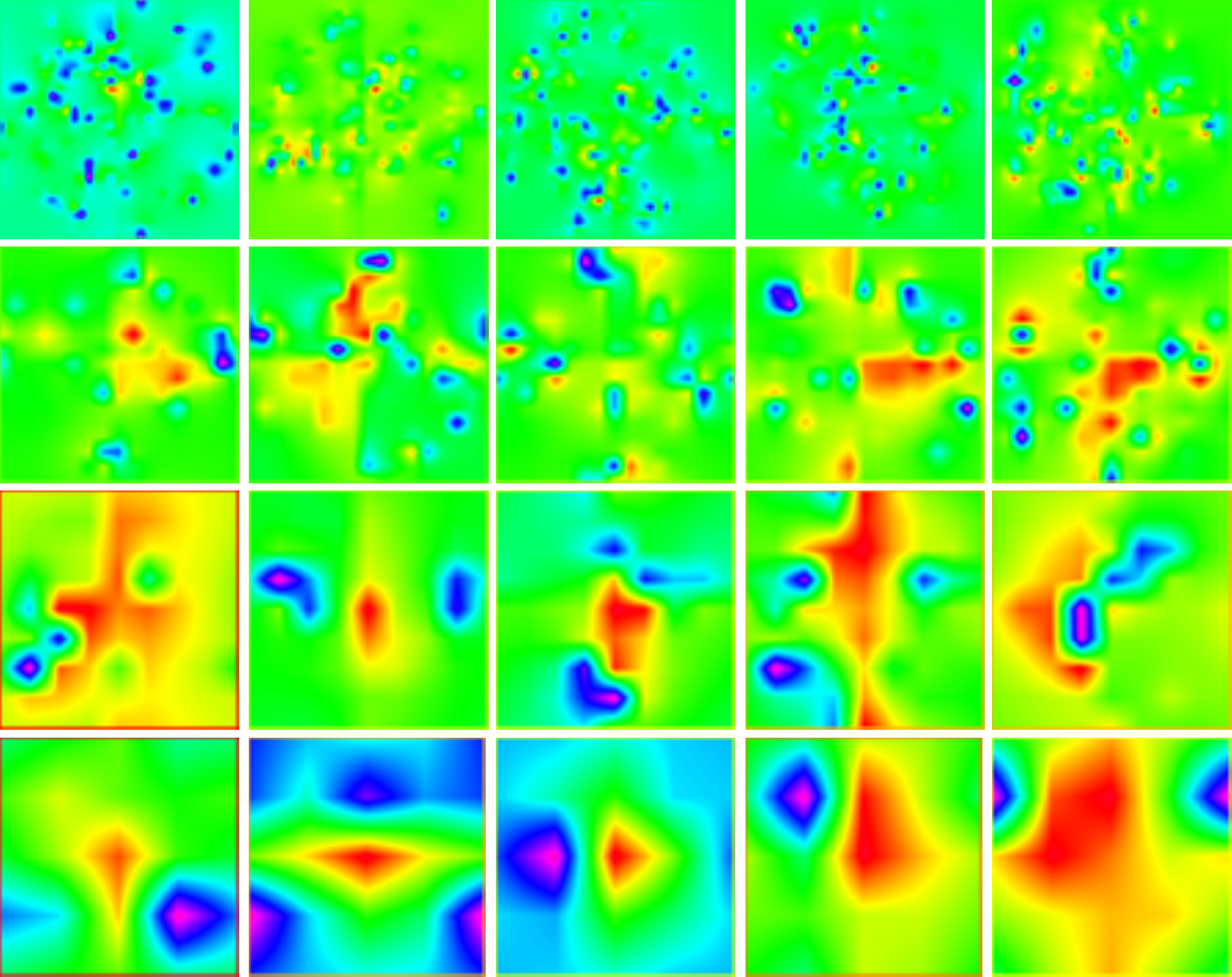}
\caption{
\label{fig:vis_emb_direct}
\textbf{Visualization of the trained embedding feature of PointNet with direct table learning (\textit{LUTI-Direct}).}
The results are from different discretizations.
From top to bottom, TV-L2 regularized \textit{LUTI-Direct}  with $D=32, 16, 8,$ and $4$.
Five randomly selected channels of the slice on the $z=0$ plane are shown.
}
\end{figure}
We investigated the embedding space trained by different architectures or different discretization\footnote{More results are shown in the supplementary (E) and (F).}.
All the visualized models were trained with the ModelNet40 classification task (\Sec\ref{subsec:analysis}).
In \Fig \ref{fig:vis_emb_mlp}, we visualize the trained embedding space of MLP with LUTI.
All LUTI variants of the network are initialized using the parameters of the naive PointNet trained for $100$ epochs.
We can observe that the embedding space evolves to achieve better performance for a given lattice size, $D$.
In \Fig \ref{fig:vis_emb_direct}, we visualize the trained embedding space that is directly trained for the table without tabulated MLP.
When $D$ is large, a large portion of the table is kept unchanged, because the error signal rarely arrives (See \Sec\ref{subsec:analysis} \textit{LUTI-Direct}).
As $D$ decreases, and the performance improves (\Fig\ref{fig:effect_D}, \textit{LUT-Direct}), the trained feature becomes smoother, and subjectively, it resembles to the ones trained by MLP or LUTI-MLP shown in \Fig \ref{fig:vis_emb_mlp}.


\section{Related Literature} 
\label{sec:related}
\prg{LUT realization of PointNet}
Utilization of the LUT for point set function was first discussed in \cite{sekikawa2019eventnet} for processing the data from an asynchronous event-based camera \cite{4444573}.
They utilized the discrete nature of the event data (the pixel location and polarity) to pre-compute the input/output relationship of the embedding MLP on LUT.
Therefore, this simple method cannot be used to process a continuous input as ours.
LUTI-MLP is easily applicable for \cite{sekikawa2019eventnet} to drastically reduce the memory footprint for the hi-res camera.

\prg{Extension of PointNet}
The extension of PointNet for a hierarchical structure is one of the active research topics for better utilization of local structures.
PointNet++ \cite{qi2017pointnetplusplus} was proposed as an improvement of PointNet, by hierarchically aggregating features in local point sets.
Another variant considers the aggregated features of nearby points with a graph-based network  \cite{dgcnn, thomas2019kpconv, wang2018dynamic, wang2018local, landrieu2018large}.
These hierarchical structures are more computationally demanding than the original structure; thus, the application of our LUTI-MLP could be beneficial.
For example, LUT-MLP can be directly used to speed up the \textit{EdgeConv} of \cite{dgcnn}.

\prg{Applications of PointNet}
The authors of  \cite{zhou2018voxelnet, lang2019pointpillars, qi2018frustum} used a variant of PointNet to detect objects from LiDAR point cloud which usually contains many more points than a simple shape, and thus, requires higher computational requirements.
The use of an aggregated point set features for point cloud registration is discussed in \cite{aoki2019pointnetlk, Sarode2019PCRNetPC, lu19,wang2019deep,gross2019alignnet, Wang2019PRNet} where an iterative evaluation of embedding may require.
LUT-MLP can be applied to a range of applications to speed up these computationally demanding embedding operations. 
\section{Conclusion and Future Work}
\label{sec:conclusion}
In this paper, we explored LUT realization of point set embedding.
With an intensive architectural analysis, we show that our LUTI-MLP speeds up the embedding for more than $80\times$ without performance degeneration.
Furthermore, the proposed formulation enables efficient analytical Jacobian computation, which has a wide range of practical applications.
Further speedup could be possible by utilizing a non-Euclidean lattice, such as a permutohedral lattice, in combination with barycentric interpolation \cite{su2018splatnet,adams2010fast}.
A future research direction would the extension of this method to hierarchical \cite{qi2017pointnetplusplus} or a graph \cite{thomas2019kpconv, wang2018dynamic, wang2018local, landrieu2018large} structure.

\newpage
\clearpage
{\small
\bibliographystyle{ieee_fullname}
\bibliography{bib}
}
\newpage
\clearpage

\renewcommand\thesection{\Alph{section}}
\renewcommand\thesubsection{\thesection.\Alph{subsection}}

\onecolumn

\begin{center}
  \textbf{\Large Tabulated MLP for Fast Point Feature Embedding}\\
  [0.2cm]
  \textbf{\Large Supplementary Material}\\
  [0.5cm]
  {\large Yusuke Sekikawa and Teppei Suzuki}\\
  [0.5cm]
  {\large DENSO IT Laboratory}\\
  [1cm]
\end{center}

\setcounter{section}{0}
\setcounter{equation}{0}
\setcounter{figure}{0}
\setcounter{table}{0}
\setcounter{page}{1}
\renewcommand{\theequation}{S\arabic{equation}}
\renewcommand{\thefigure}{S\arabic{figure}}
\renewcommand{\thetable}{S\arabic{table}}

\section{T-Net for Feature Transformation}
\begin{figure}[!h]
\includegraphics[width=1\columnwidth]{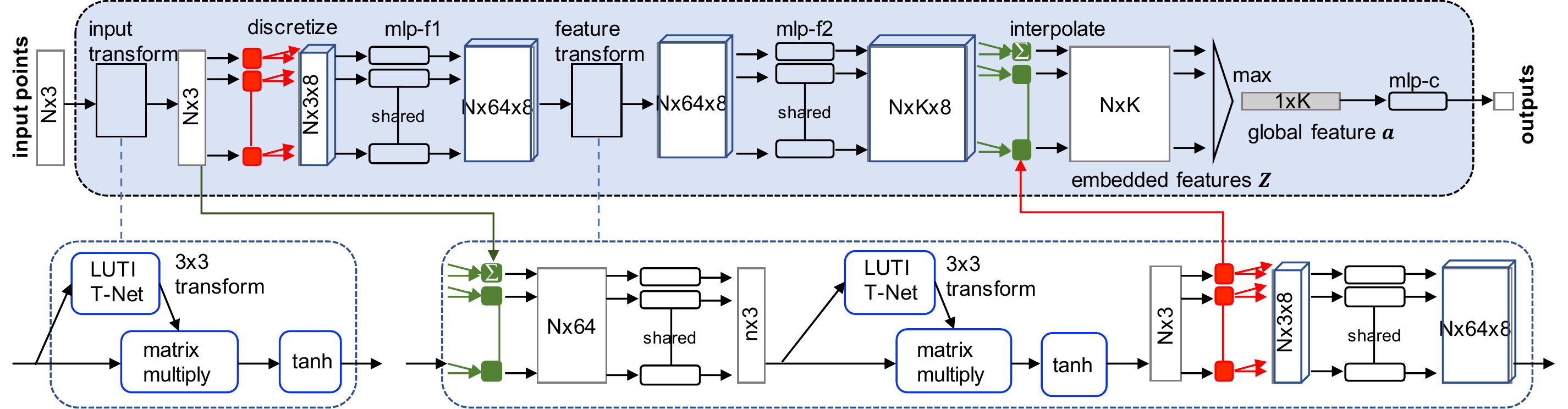}
\caption{
\textbf{Detailed network architecture of PointNet with LUTI-MLP (at training).}
As discussed in the main paper,  we modified the original T-Net (IT) by adding $\tanh$ function to bound the output transformed coordinates.
T-Net for FT further includes two linear layers coupled with interpolation/discretization module that decrease and increase the input feature, to make the dimension of  LUT to a manageable lattice size  ($\mathbb{D}^3$).
\label{fig:tnet}
}
\end{figure}
In the main paper, we focused our discussion on PointNet with and without T-Net for input transformation (IT) and skipped the discussion about T-Net for feature transformation (FT).
Original T-Net (FT) consumes $N\times64$ vector to estimate  $64\times 64$ affine matrix.
It is hard to precompute and store MLP on the 64-dimensional lattice ($\mathbb{D}^{64}$), so we modified T-Net (FT) so that the dimension of lattice becomes reasonably small  ($\mathbb{D}^{3}$) as shown in \Fig \ref{fig:tnet}.
Our modified T-Net (FT) includes two linear layers coupled with the interpolation/discretization module that decrease and increase the input feature, to make the dimension of  LUT to a manageable size ($\mathbb{D}^3$).
We did not use regularization, which encourages the matrix to be orthogonal,  for the modified T-Net (FT) because it has fewer parameters (same as T-Net (IT)) than the original $64\times64$ affine matrix.
\Tab\ref{table:modelnet_ft} shows the results from PointNet with T-Net (FT) on ModelNet40 \cite{wu20153d} shape classification benchmark.
Similar to the results on the main paper, E2E trained PointNet with our LUTI-MLP achieved comparable performance with its counterpart using MLP.
At test time, all the embedding MLP on the main network, T-Net (IT), and T-Net (FT) are realized as LUTI.
\begin{table}[h]
\caption{
\label{table:modelnet_ft}
\textbf{Results with T-Net (FT)  on ModelNet40 \cite{wu20153d} classification task.}
For all the results, T-Net (FT) is used in combination with T-Net (IT).
We compare the results from PointNet with MLP and one with our LUTI-MLP.
MLP IT+FT represents the original PointNet with IT+FT (with orthogonal regularization).
IT+FT (mod) represents modified version of T-Net shown in \Fig \ref{fig:tnet}.
Results using different discretization $D$ are shown for LUTI-MLP, and the result using MLP are shown in column $\inf$.
Our E2E trained PointNet with LUT-MLP achieves comparable results as original architecture while achieving significant speed-up.
}

\begin{center}
\begin{tabular}{cc|cccccc}
\multicolumn{2}{c}{  } & \multicolumn{6}{c}{discretization $D$} \\ \hdashline
\multicolumn{2}{c}{ architecture } & $\inf$ & 64 & 32 & 16 & 8 & 4 \\ \hline
MLP & IT+FT & 86.47\% & - & - & - & - & - \\ \hdashline
MLP & IT+FT(mod) & 85.74\% & - & - & - & - & - \\ \hdashline
LUT-MLP Approx. & IT+FT(mod) & - & 85.45\% & 84.28\% & 80.92\% & 42.63\% & 11.18\% \\
LUTI-MLP Approx. & IT+FT(mod) & - & 84.52\% & 80.71\% & 54.78\% & 19.08\% & 9.89\% \\ \hdashline
LUTI-MLP E2E & IT+FT(mod) & - & 85.37\% & 85.58\% & 84.77\% & 86.02\% & 84.77\% \\
\end{tabular}

\end{center}
\end{table}

\newpage
\section{Parts Segmentation Results with Different Discretization}
\begin{figure}[!h]
\includegraphics[width=1\columnwidth]{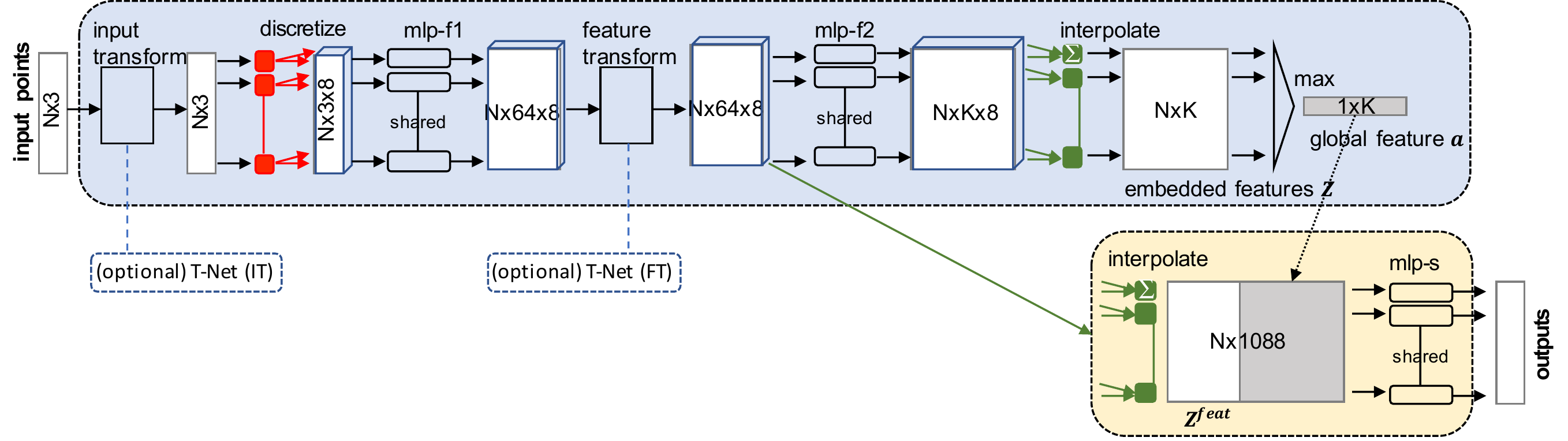}
\caption{
\textbf{PointNet for point-wise segmentation with LUTI (at training).}
Other than the embedded MLP is replaced with our LUTI-MLP, the rest of the network architecture is the same as the point-wise segmentation network of PointNet \cite{qi2017pointnet}.
At test time both $\mathbf{Z}$ and $\mathbf{Z}^{feat}$ are computed as a trilinear interpolation of LUT sized $D\times D\times D\times (1024+64)$
\label{fig:network_seg}
}
\end{figure}
\Tab \ref{table:shapenet_seg} shows additional results of parts segmentation task using ShapeNet \cite{shapenet2015}, evaluated with different discretization  $D$ that was omitted from the main paper (\Sec\ref{subsec:exp_app}, 3D Object Part Segmentation, \Tab\ref{table:shapenet_digest}). 
The performance is equally good across different discretizations.
These results infer that our LUTI-MLP, even with small size lattice, is also effective for point-wise estimation network (\Fig\ref{fig:network_seg}), which is largely different from the classification network (\Fig\ref{fig:overview}, bottom).
\begin{table}[h]
\caption{
\label{table:shapenet_seg}
$\textbf{Segmentation results on ShapeNet part dataset.}$ 
Our PointNet with LUT-MLP achieves comparable results as original PointNet with MLP on most of the ranges of $D$ while achieving significant speed-up.
Metric is mIoU(\%) on points. 
IT represent T-Net (IT).
}

\begin{center}
\begin{tabular}{ccc|c|cccc}
& IT & D & mean & chair & table & aero & lamp \\ \hdashline
\# shapes & & & & 3758 & 5271 & 2690 & 1547 \\\hline
MLP & & - & 84.73 & 88.09 & 92.59 & 79.80 & 78.44 \\
MLP & \checkmark & - & 84.38 & 88.09 & 91.32 & 76.19 & 81.90 \\ \hdashline
LUTI & & 32 & 85.38 & 88.72 & 92.84 & 78.67 & 81.29 \\
LUTI & \checkmark & 32 & 84.33 & 88.12 & 91.61 & 76.49 & 81.08 \\ \hdashline
LUTI & & 16 & 85.42 & 88.62 & 91.67 & 81.38 & 80.01 \\
LUTI & \checkmark & 16 & 85.66 & 88.86 & 91.98 & 80.29 & 81.50 \\ \hdashline
LUTI & & 8 & 85.56 & 88.58 & 92.54 & 79.28 & 81.83 \\
LUTI & \checkmark & 8 & 85.36 & 87.89 & 91.52 & 82.01 & 80.03 \\ \hdashline
LUTI & & 4 & 84.69 & 87.95 & 92.48 & 80.07 & 78.28 \\
LUTI & \checkmark & 4 & 84.41 & 87.39 & 91.76 & 78.78 & 79.71 \\
\end{tabular}
\end{center}
\end{table}

\newpage
\section{Point Set Registration Accuracy of PointNetLK with LUTI}
\begin{figure}[!h]
\includegraphics[width=1\columnwidth]{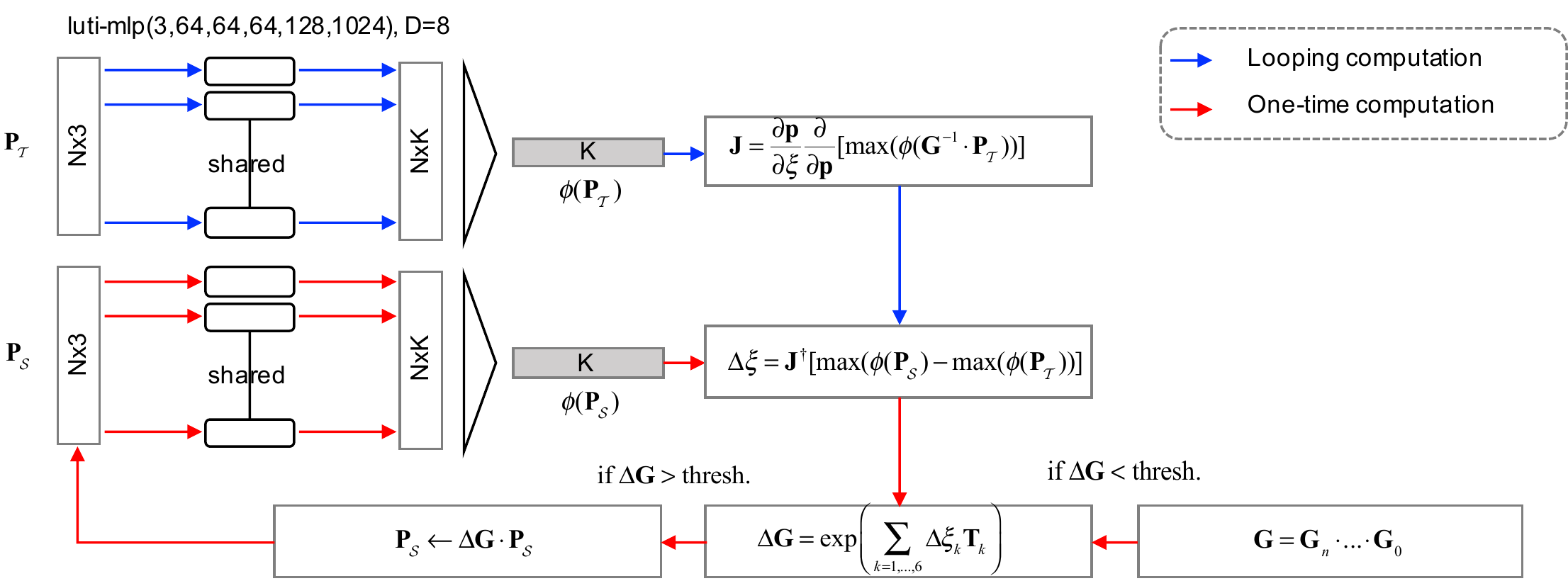}
\caption{
\textbf{Network architecture of PointNetLK with "LUTI-MLP embedding" and "LUTI-MLP canonical Jacobian" (at test time).}
It is used for point-set registration tasks.
Other than the embedded MLP is replaced with our LUTI-MLP and Jacobian computation is replace with canonical Jacobian using LUTI-MLP, the rest of the network architecture is the same as the original PointNetLK \cite{aoki2019pointnetlk}.
\label{fig:network_PNLK}
}
\end{figure}
\begin{figure}[!h]
\includegraphics[width=0.49\columnwidth]{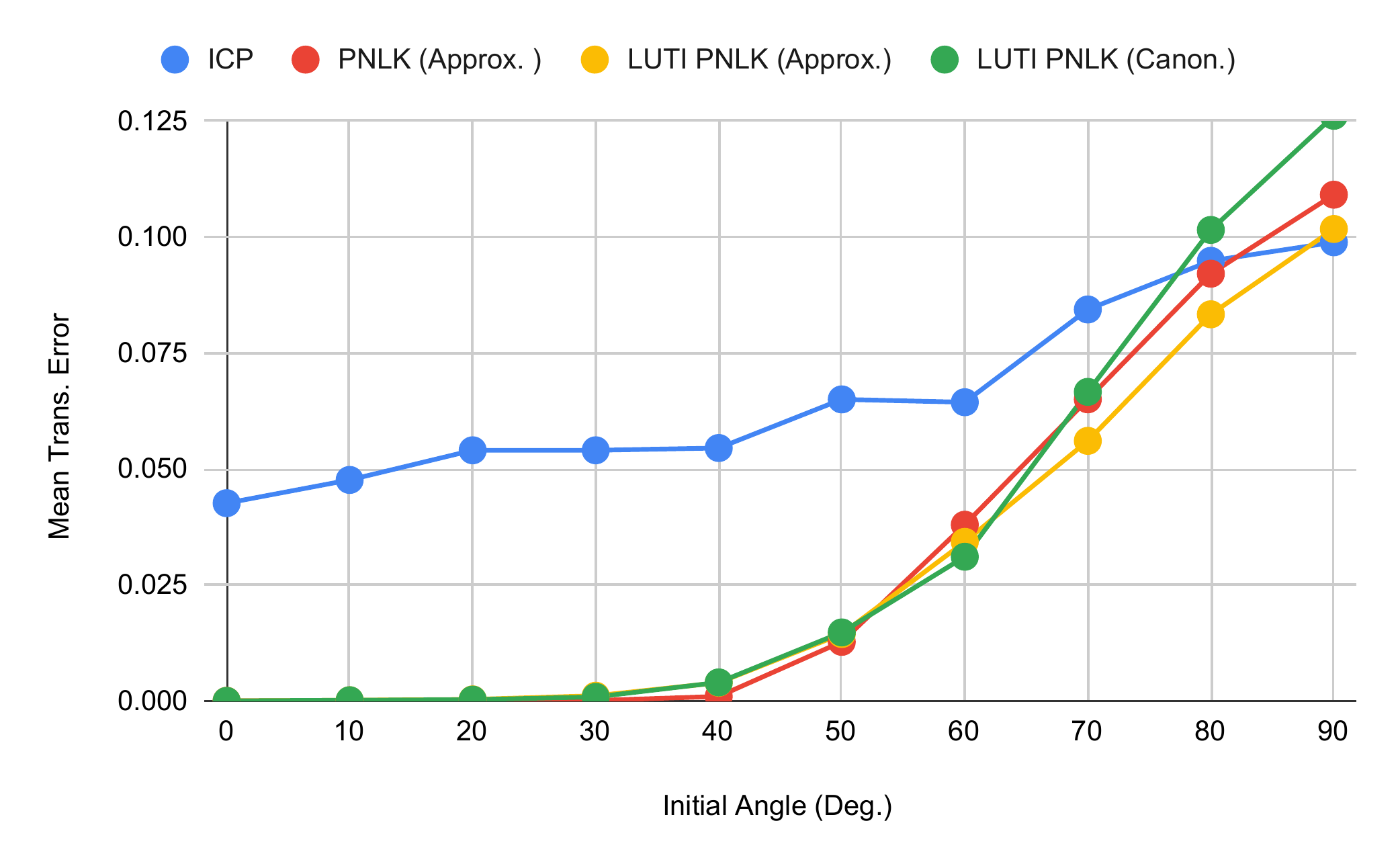}
\includegraphics[width=0.49\columnwidth]{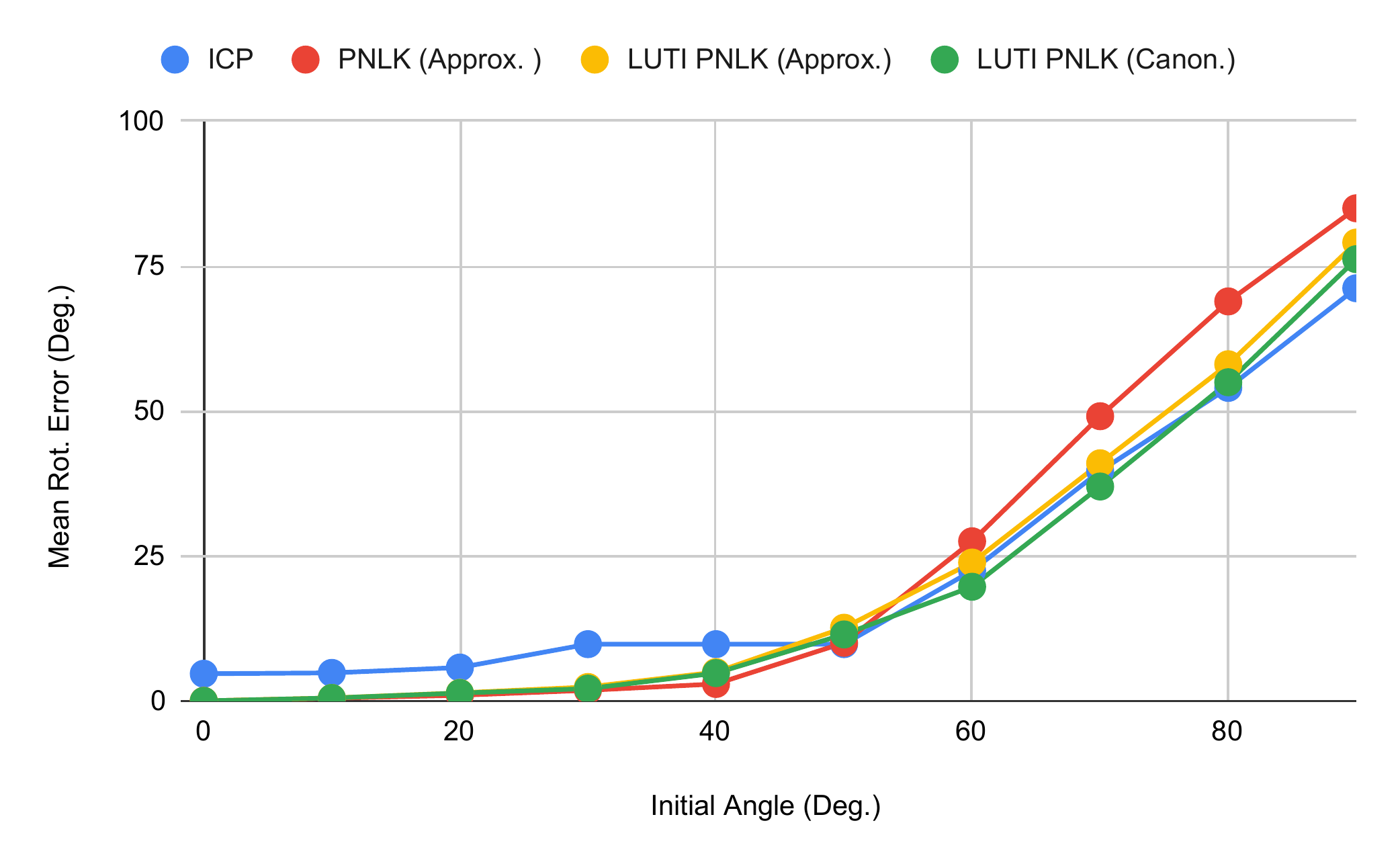}
\caption{
\label{fig:PTLK}
\textbf{Results for point cloud registration with PointNetLK with LUTI-MLP.}
The results compare the performance of original PointNetLK, the variant using LUTI for approximate Jacobian, and variant using LUTI for canonical Jacobian.
Results are reported for 10 iterations for all the architecture.
The performance is evaluated on categories unseen during training.
Both of LUTI integrated networks achieve comparable performance while achieving about $37\times$ the speeds of the overall registration process.
}
\end{figure}
\Fig\ref{fig:network_PNLK} shows the network architecture of PointNetLK with "LUTI-MLP embedding" and "LUTI-MLP canonical Jacobian" (at test time).
\Fig \ref{fig:PTLK} shows accuracy from point cloud registration task discussed in experiments (\Sec\ref{subsec:exp_app}, 3D Point Cloud Registration) using ModelNet40 \cite{wu20153d}.
As shown in \Fig\ref{fig:PTLK}, the registration accuracy of the three variants, finite-difference of MLP \cite{aoki2019pointnetlk}, finite-difference of LUTI-MLP (ours), and canonical Jacobian of LUTI-MLP  (ours) are almost the same.
For this experiment, we followed the protocol of PointNetLK \cite{aoki2019pointnetlk} by using their published source code\footnote{https://github.com/hmgoforth/PointNetLK} and modified feature embedding part and Jacobian computation part with our LUTI-MLP (\Fig\ref{fig:network_PNLK}).
We did not evaluate the registration accuracy of the variant using canonical Jacobian of MLP; it is because computational times for training and testing of this variant are computational demanding (see \Tab\ref{table:speed_jac}).

\newpage
\section{Training PointNet with LUTI-MLP from Scratch}
The results of E2E trained PointNet with LUTI-MLP (\textit{LUTI-MLP E2E}) in the main paper (\Fig\ref{fig:effect_D}) used pre-trained model (at $100$ epochs) from vanilla PointNet (with MLP) for initializing the network parameters for training.
The primary purpose of the pre-training was to see the evolution of the embedding feature by the E2E training (\Sec \ref{subsec:exp_vis}). 
\Tab\ref{tab:scrach_vs_pretrain} shows the results from the PointNet with LUTI-MLP trained from scratch without the pre-training.
For all ranges of $D$ in our experiments, the performance was comparable with the ones from the pre-trained network.
As this table infer, this pre-training has little impact on classification accuracy.
\begin{table}[h]
\caption{
\label{tab:scrach_vs_pretrain}
\textbf{Comparison of object classification accuracy of PointNet using LUTI-MLP  with and without pre-training. }
The classification accuracy of the network without pre-training is comparable with the pre-trained network. The results from differenct discretization $D$ are shown.
}

\begin{center}
\begin{tabular}{ccc|ccccc}
\multicolumn{3}{c}{ Algorithm } & \multicolumn{5}{c}{ D } \\
& & IT & 64 & 32 & 16 & 8 & 4 \\ 
\hline
\multirow{2}{*}{ pre-train } & LUTI MLP E2E & & 85.86\% & 85.09\% & 86.22\% & 85.29\% & 85.37\% \\
& LUTI MLP E2E & \checkmark & 88.09\% & 87.91\% & 87.24\% & 87.76\% & 87.24\% \\
\\ \hdashline
\multirow{2}{*}{ scratch } & LUTI MLP E2E & & 85.90\% & 85.37\% & 86.51\% & 86.26\% & 85.37\% \\
& LUTI MLP E2E & \checkmark & 86.83\% & 86.67\% & 87.52\% & 87.60\% & 86.87\% \\
\end{tabular}

\end{center}
\end{table}

\newpage
\section{Visualization of Embedding Space of \textit{LUTI-MLP E2E}}
\begin{figure}[!h]
\includegraphics[width=1\columnwidth]{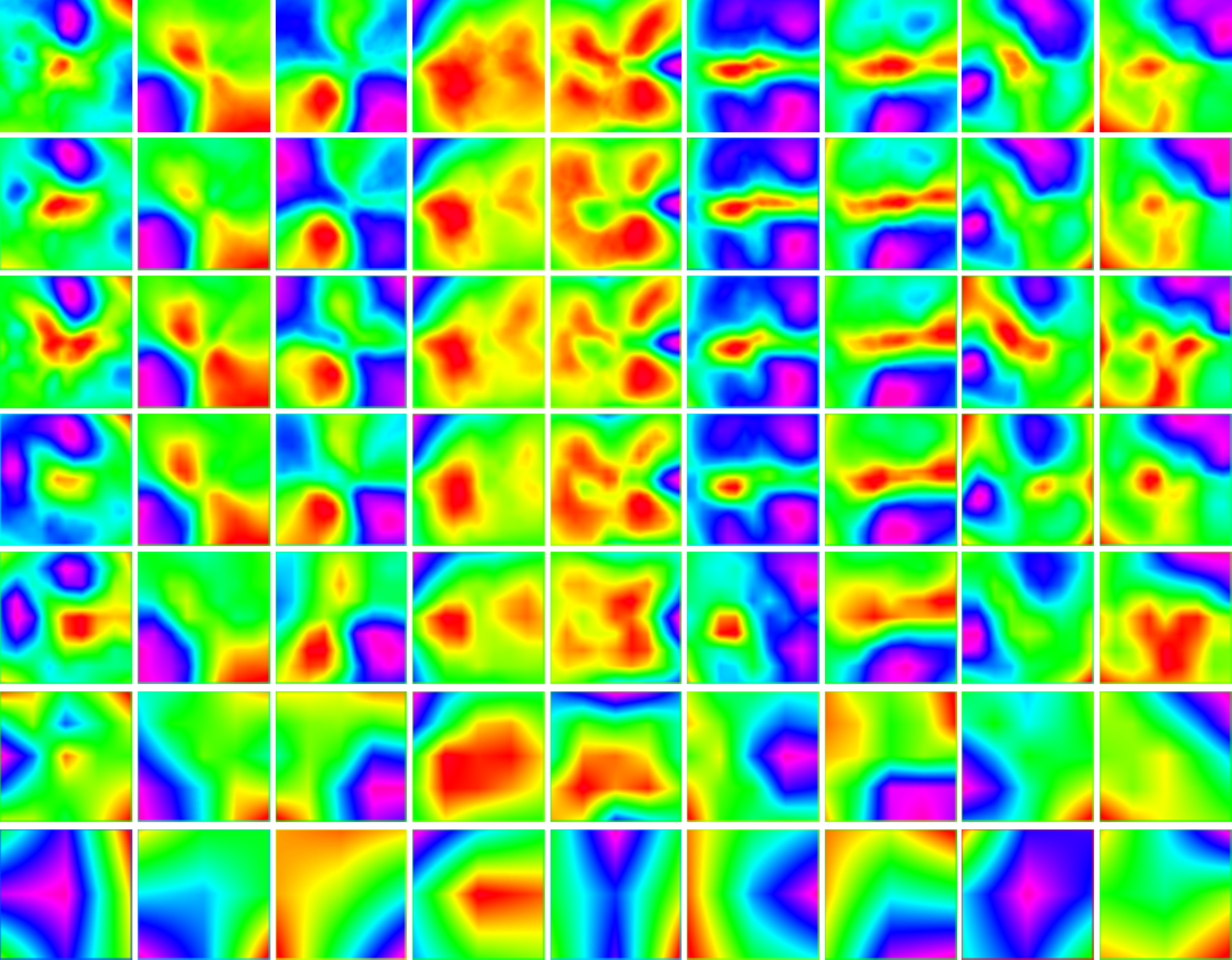}
\caption{
\label{fig:vis_emb_mlp_supp}
\textbf{Visualization of the trained embedding feature of PointNet using MLP (\textit{MLP}) and PointNet with LUTI-MLP (\textit{LUTI-MLP E2E}).}
The results are reported for different discretizations $D$  for \textit{LUTI-MLP}.
From top to bottom, \textit{MLP} (PointNet), \textit{LUTI-MLP} with $D=64,32, 16,8, 4$ and $2$.
Note that the  LUTI-MLP use pre-trained MLP from naive PointNet (\textit{MLP} on \Sec \ref{subsec:analysis}).
Nine randomly selected channels of the slice on the $z=0$ plane are shown.
}
\end{figure}

More results of trained embedding feature space from \textit{LUTI-MLP E2E} (\Fig \ref{fig:vis_emb_mlp} in the main paper) are shown in \Fig \ref{fig:vis_emb_mlp_supp} for more more diverse discretization $D$ and channel.

\newpage
\section{Visualization of Embedding Space from Direct Table Leaning (\textit{LUTI-Direct})}
Trained embedding space of direct table training are shown in \Fig \ref{fig:vis_emb_direct_supp}.
In this figure, we also visualize the results from diffident regularization.
Compared with the variants trained with MLP (\Fig \ref{fig:vis_emb_mlp_supp}, top), the learned embedding feature of the variant tend to have more peaks when the discretization is fine.

As discussed in the main paper, we used regularization using the total variation (TV)  with $p$ norm (TVL1 for $p=1$, TVL2 for $p=2$) for parameter (table) $\mathbf{W}$. 
The TV was evaluated using  3D spatial neighbor $\mathcal{N}$  as
\begin{equation}
T V(\mathbf{W})=\sum_{i, k \in \mathcal{N}}\left\|\mathbf{w}_{i}-\mathbf{w}_{j}\right\|^{p}.
\end{equation}
It was used in combination with standard classification objective $l_{CE}$.
Then, the loss function becomes,
\begin{equation}
l=l_{CE} + \lambda T V(\mathbf{W}),
\end{equation}
where we used $\lambda=1.0$ for experiments in  \Sec \ref{subsec:analysis} (\textit{LUT-Direct}  / \textit{LUTI-Direct}) using regularization.
Results from using regularization TVL1 or TVL2 tend to be more smooth than one without it.
\begin{figure}[!h]
\begin{subfigure}[t]{1.0\columnwidth}
\includegraphics[width=1\columnwidth]{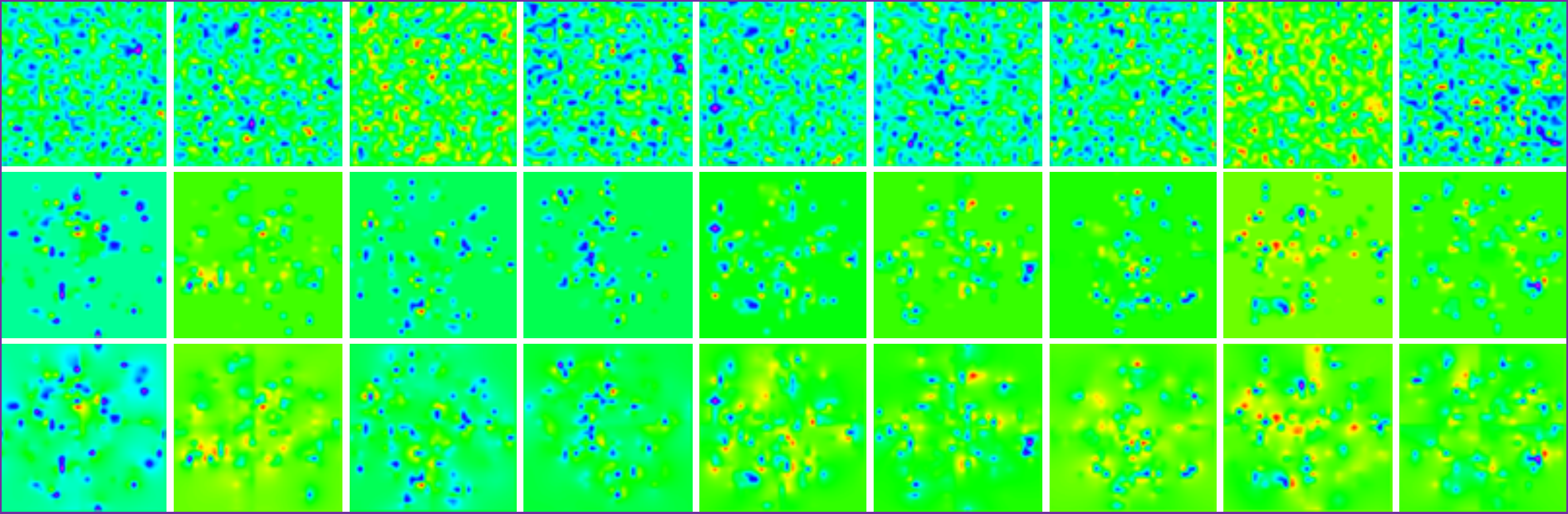}
\subcaption{$D=32$}
\end{subfigure}

\begin{subfigure}[t]{1.0\columnwidth}
\includegraphics[width=1\columnwidth]{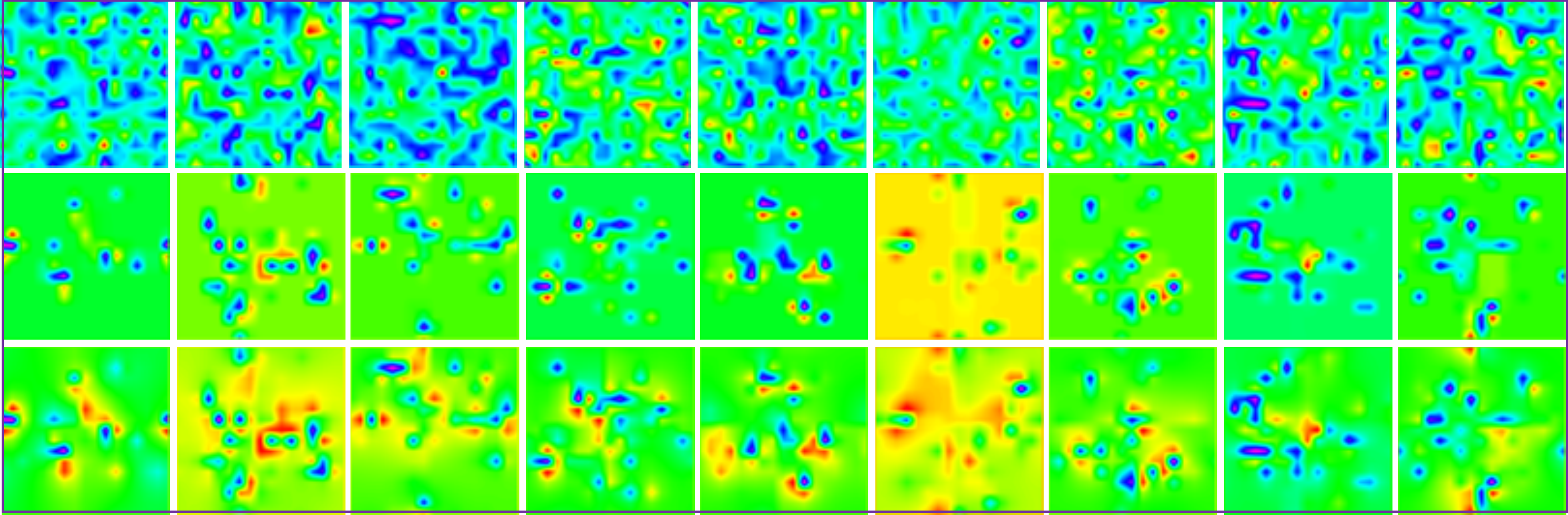}
\subcaption{$D=16$}
\end{subfigure}
\end{figure}
\clearpage

\begin{figure}[!h]
\begin{subfigure}[t]{1.0\columnwidth}\ContinuedFloat
\includegraphics[width=1\columnwidth]{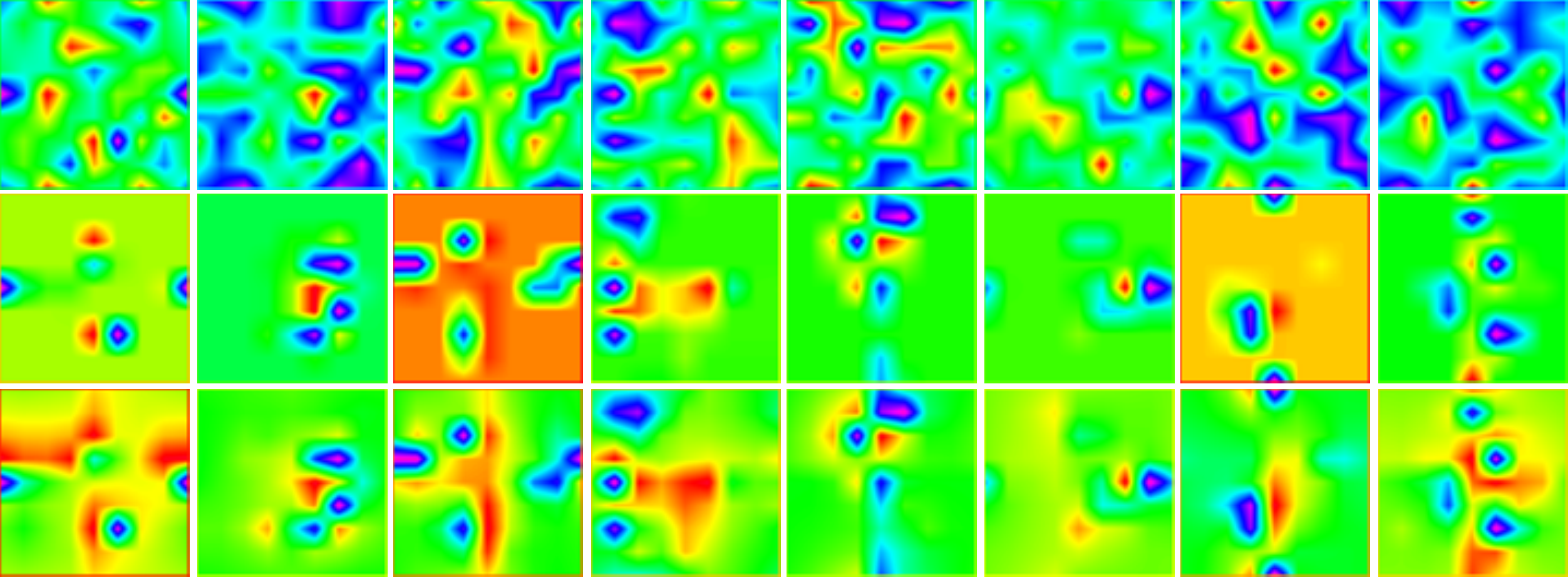}
\subcaption{$D=8$}
\end{subfigure}

\begin{subfigure}[t]{1.0\columnwidth}
\includegraphics[width=1\columnwidth]{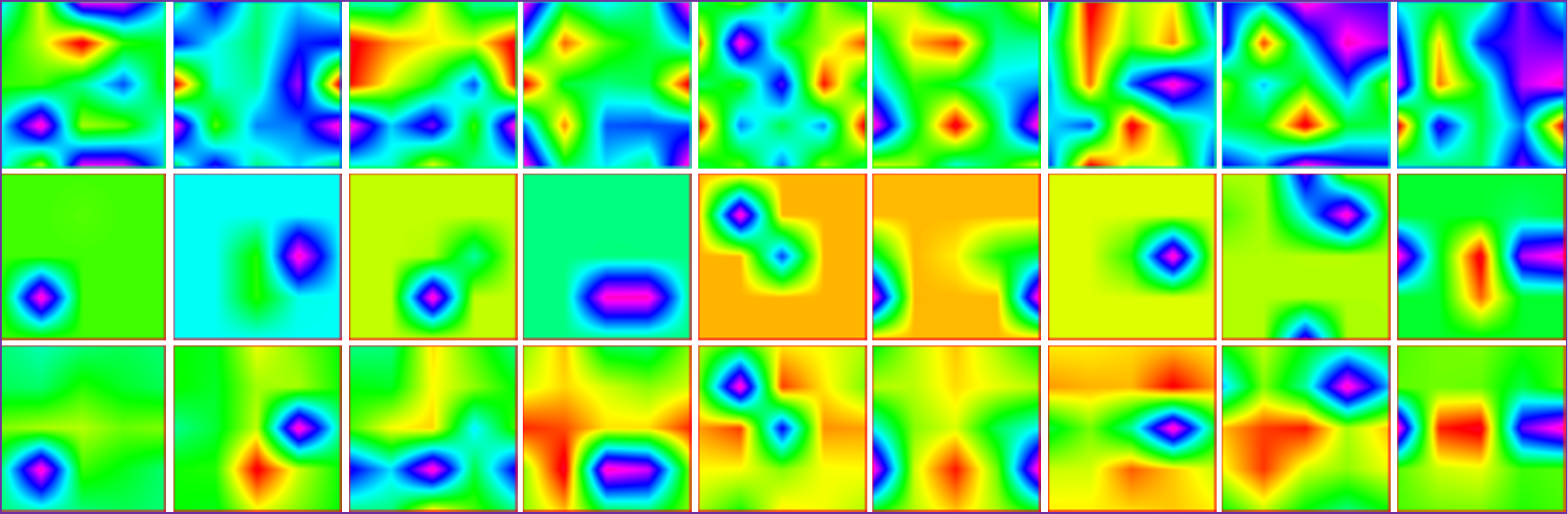}
\subcaption{$D=4$}
\end{subfigure}

\begin{subfigure}[t]{1.0\columnwidth}
\includegraphics[width=1\columnwidth]{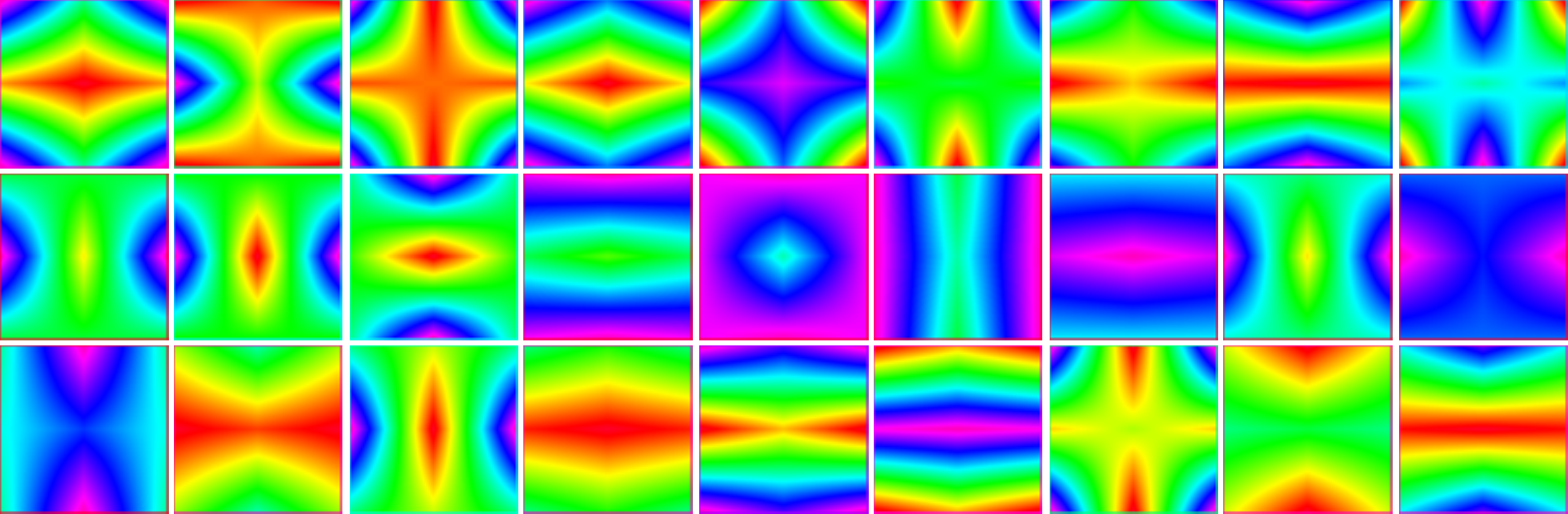}
\subcaption{$D=2$}
\end{subfigure}
\caption{
\label{fig:vis_emb_direct_supp}
\textbf{Visualization of the trained embedding feature of PointNet with direct table learning (\textit{LUTI-Direct}).}
The results are from different discretization $D$ and different regularization.
From top to bottom, \textit{LUTI-Direct}  with $D=32, 16, 8, 4$ and $2$.
For each group in the same discretization,  without regularization (Top),  TVL1 regularization (Middle), and TV L2 regularization (Bottom).
Nine randomly selected channels of the slice on the $z=0$ plane are shown.
}
\end{figure}

\section{Memory Footprint Consideration}
\Tab\ref{table:memory}  shows the memory footprint comparison between different lattice $\mathbb{D}^M$ with discretization $D$ and lattice dimension $M$.
When $D=4$ with $N=3$, the size of the table is about 250 KB, which is even smaller ($2.3\times$) than that of the parameter of MLP, which is used as the embedding function (\Fig\ref{fig:overview}, top) of original PointNet \cite{qi2017pointnet}.
In the main paper, we focused our discussion on 3D data ($\mathbb{R}^3$) such as point cloud data; thus, the size of the lattice is $\mathbb{D}^3$.
Our LUT-MLP can be extensible to  $\mathbb{R}^M$ using $\mathbb{D}^M$ lattice as long as a memory for LUT permits.
When we want to incorporate additional features, such as intensity, color, normal, etc., the lattice dimension $M$ needs to be extended as well, which increases the memory footprint exponentially, as shown in \Tab\ref{table:memory}.
For example, when we incorporate the intensity information from LiDAR data (the input point feature dimension would be $\mathbb{R}^4$), and the lattice becomes $\mathbb{D}^4$.
In this case, about $16$MB of memory to store the LUT is required for $D=8$. 
When the normal information is further incorporated (the input point feature dimension would be $\mathbb{R}^6$),  it requires about $1$ gigabyte of memory for $D=8$.
\begin{table}[h]
\caption{\label{table:memory}
\textbf{Comparison of the memory footprint of MLP and LUTI-MLP.}
We compare the memory footprint (MB) of LUTI for different discretization $D$ and different lattice dimension $M$ for $N$-D lattice $\mathbb{D}^M$ with MLP.
A $4$ byte floating point is assumed for storing each parameter.
}

\begin{center}
\begin{tabular}{c|c:cccccccc}
& MLP & \multicolumn{5}{c}{ LUTI with D } & & & \\ 
\hline
M & & 1024 & 64 & 32 & 16 & 8 & 5 & 4 & 2 \\ \hdashline
3 & 5.68E-01 & 4.19E+06 & 1.02E+03 & 1.28E+02 & 1.60E+01 & 2.00E+00 & 4.88E-01 & 2.50E-01 & 3.13E-02 \\
4 & 5.68E-01 & 4.29E+09 & 6.55E+04 & 4.10E+03 & 2.56E+02 & 1.60E+01 & 2.44E+00 & 1.00E+00 & 6.25E-02 \\
5 & 5.69E-01 & 4.40E+12 & 4.19E+06 & 1.31E+05 & 4.10E+03 & 1.28E+02 & 1.22E+01 & 4.00E+00 & 1.25E-01 \\
6 & 5.69E-01 & 4.50E+15 & 2.68E+08 & 4.19E+06 & 6.55E+04 & 1.02E+03 & 6.10E+01 & 1.60E+01 & 2.50E-01 \\
\end{tabular}

\end{center}
\end{table}

\end{document}